\documentclass{article} 
\usepackage{iclr2027_conference,times}

\usepackage{amsmath,amsfonts,bm}

\def\eqref#1{equation~\ref{#1}}

\def\1{\bm{1}}

\DeclareMathAlphabet{\mathsfit}{\encodingdefault}{\sfdefault}{m}{sl}
\SetMathAlphabet{\mathsfit}{bold}{\encodingdefault}{\sfdefault}{bx}{n}

\usepackage{amsmath}
\usepackage{amssymb}
\usepackage{graphicx}
\usepackage{booktabs}
\usepackage{multirow}
\usepackage{subcaption}
\usepackage{float}
\usepackage{placeins}

\usepackage{hyperref}
\usepackage{url}

\title{BeatFlow-ECG: Rectified Flow for ECG Reconstruction from Indirect Wearable Signals}

\author{
  \textbf{Mohamed Kamel}$^{1}$\thanks{Corresponding author: \texttt{kamelms@vcu.edu}}, \textbf{Sahar Selim}$^{2}$, \textbf{Walaa Medhat}$^{2}$, \textbf{Tamer Nadeem}$^{1}$ \\
  $^{1}$Department of Computer Science, Virginia Commonwealth University, Richmond, VA, USA \\
  $^{2}$School of Information Technology and Computer Science, Nile University, Giza, Egypt
}

\iclrfinalcopy
\begin{document}

\maketitle

\begin{abstract}
Continuous cardiac monitoring outside clinical settings requires signals that are both informative and practical to collect during daily life. Electrocardiography (ECG) provides rich information about cardiac rhythm and waveform morphology, while wearable photoplethysmography (PPG) is easier to acquire continuously but is only an indirect cardiovascular measurement and is highly sensitive to motion. We present \textbf{BeatFlow-ECG}, a conditional rectified-flow model for reconstructing single-channel ECG from synchronized PPG and inertial measurements. BeatFlow-ECG models reconstruction as conditional transport from noise to ECG using a convolutional encoder-decoder with a transformer bottleneck and explicit flow-time conditioning. Motion information is incorporated through IMU-derived conditioning features, motion-dependent loss weighting, and an easy-to-hard training curriculum. We evaluate the model under leave-one-subject-out protocols on PPG-DaLiA and WESAD. BeatFlow-ECG achieves the best results among the evaluated deterministic, adversarial, and diffusion-based baselines across all reported waveform and beat-timing metrics, with Pearson correlations of 0.983 and 0.986 and R-peak F1 scores of 0.946 and 0.955, respectively. Compared with Conditional DDPM-1D, L1 error decreases from 0.085 to 0.062 on PPG-DaLiA and from 0.074 to 0.055 on WESAD. Additional analyses on PPG-DaLiA show higher correlation in fixed R-peak-relative waveform regions and lower reconstruction error across low-, medium-, and high-motion subsets.
\end{abstract}

\section{Introduction}

Electrocardiography (ECG) captures cardiac electrical activity with high temporal precision and provides clinically useful information about rhythm and waveform morphology. However, continuous ECG acquisition outside clinical settings typically requires dedicated electrodes and specialized hardware. Wearable photoplethysmography (PPG) is easier to acquire continuously, but it provides only an indirect measure of cardiovascular activity and is sensitive to motion artifacts and signal degradation \citep{Allen2007-vn,electronics3020282,Dias2018-kq,Elgendi2024-iq}. Reconstructing ECG from wearable signals could support ambulatory cardiac monitoring, but the relationship between peripheral pulse measurements and ECG varies across subjects and activity conditions \citep{ppg-dalia_495,s19143079}. Accurate reconstruction therefore requires preserving both local waveform morphology and longer-range beat timing.

Prior work has explored ECG reconstruction and restoration using deterministic, adversarial, and diffusion-based models \citep{8834612,9506990,Sarkar_Etemad_2021,shome2023regiondisentangleddiffusionmodelhighfidelity,Ezzat2024-rs,s24186046,NEURIPS2024_9988f2c8}. However, most existing work does not address subject-independent reconstruction from multimodal wearable signals under substantial motion. Diffusion-based methods also require iterative sampling, which can increase inference cost \citep{ho2020denoisingdiffusionprobabilisticmodels,song2021scorebasedgenerativemodelingstochastic,song2022denoisingdiffusionimplicitmodels}.

We present \textbf{BeatFlow-ECG}, a conditional rectified-flow framework for reconstructing single-channel ECG from synchronized PPG and IMU signals. Instead of directly regressing the waveform in one step, BeatFlow-ECG models reconstruction as conditional transport from noise to ECG. The model combines a convolutional encoder-decoder with a transformer bottleneck and explicit flow-time conditioning. Motion information is incorporated through IMU-derived conditioning features, motion-dependent loss weighting, and an easy-to-hard training curriculum.

We evaluate BeatFlow-ECG on PPG-DaLiA \citep{ppg-dalia_495} and WESAD \citep{10.1145/3242969.3242985} under leave-one-subject-out (LOSO) protocols.

Our main contributions are:
\begin{itemize}
    \item We formulate single-channel ECG reconstruction from synchronized wearable PPG and inertial measurements as a conditional rectified-flow problem, modeling reconstruction as continuous transport from noise to ECG.
    \item We incorporate motion information through IMU-derived conditioning features, motion-weighted velocity matching, and a curriculum that gradually introduces higher-motion windows, and evaluate the contribution of these components through ablations.
    \item We evaluate BeatFlow-ECG under LOSO on PPG-DaLiA and WESAD against deterministic, capacity-matched, adversarial, and diffusion-based baselines. BeatFlow-ECG achieves the best overall waveform and beat-timing results among the evaluated models, with additional analyses examining regional waveform agreement, motion robustness, and inference efficiency.
\end{itemize}

\section{Related Work}

\subsection{ECG reconstruction from indirect wearable signals}

ECG reconstruction from indirect physiological signals has been studied using several approaches. Early work mapped photoplethysmography (PPG) to ECG \citep{8834612,9506990}, while later studies considered dictionary-learning formulations \citep{9054242} and neural models for specific ECG components such as the QRS complex \citep{9109576}. More recent methods include GAN-based synthesis \citep{Sarkar_Etemad_2021}, attention-based reconstruction \citep{Ezzat2024-rs}, modified U-Net architectures \citep{s24186046}, and diffusion-based translation \citep{shome2023regiondisentangleddiffusionmodelhighfidelity}. Most of these methods use PPG alone and do not explicitly use synchronized IMU measurements to account for motion. Reported results also differ in datasets, preprocessing, train/test splits, signal lengths, sampling rates, and evaluation protocols, making direct numerical comparison across papers difficult. BeatFlow-ECG instead combines wrist PPG with IMU-derived motion information and is evaluated under subject-independent LOSO. We also evaluate CardioGAN under the same LOSO protocol on both PPG-DaLiA and WESAD.

\subsection{ECG restoration, morphology reconstruction, and lead reconstruction}

ECG restoration has also been studied when ECG is available in partial, degraded, or reduced-dimensional form. Representative directions include ECG denoising \citep{8802155,8693790,9018175,Li_2024}, imputation and restoration \citep{xu2023pulseimputenovelbenchmarktask,jenkins2023improvingdiffusionmodelsecg}, and lead reconstruction from reduced or single-lead observations \citep{10191213,9789045,ZHAN2024106377,10.1007/978-3-031-43990-2_18,10.1007/978-3-031-70378-2_17,10.1145/3730436.3730464,s26010212}. BeatDiff \citep{NEURIPS2024_9988f2c8}, which is closely related to our work, uses diffusion for ECG beat morphology reconstruction from indirect ECG observations. These methods start from ECG observations and restore missing or corrupted ECG content. In contrast, BeatFlow-ECG reconstructs ECG directly from peripheral PPG and IMU measurements.

\subsection{Generative modeling foundations for physiological time series}

Diffusion probabilistic models \citep{ho2020denoisingdiffusionprobabilisticmodels,song2021scorebasedgenerativemodelingstochastic,song2022denoisingdiffusionimplicitmodels} and their conditional and inverse-problem extensions \citep{10.5555/3600270.3601984,chung2024diffusionposteriorsamplinggeneral,kawar2021snipssolvingnoisyinverse,debortoli2023diffusionschrodingerbridgeapplications,zhou2023denoisingdiffusionbridgemodels} have been widely studied for structured generation. In physiological signals, diffusion models have been applied to synthetic ECG generation, conditional ECG modeling, and multimodal cardiovascular signal generation \citep{adib2023syntheticecgsignalgeneration,alcaraz2023diffusionbasedconditionalecggeneration,Lai_2025,LIN2025104948,chen2025versatilecardiovascularsignalgeneration}. Diffusion-based methods typically rely on iterative sampling, which can increase inference cost \citep{ho2020denoisingdiffusionprobabilisticmodels,song2021scorebasedgenerativemodelingstochastic,song2022denoisingdiffusionimplicitmodels}. Rectified flow and flow-matching methods provide continuous-time alternatives \citep{lipman2023flowmatchinggenerativemodeling,liu2022flowstraightfastlearning,albergo2025stochasticinterpolantsunifyingframework,esser2024scalingrectifiedflowtransformers,zhang2025trajectoryflowmatchingapplications}. BeatFlow-ECG applies this formulation to wearable ECG reconstruction with PPG/IMU conditioning and motion-aware training.

\section{Method}
We present \textbf{BeatFlow-ECG}, a conditional rectified-flow model for reconstructing ECG from wearable measurements. We describe the rectified-flow formulation, model architecture, motion-aware training, and preprocessing pipeline below.

\subsection{Problem formulation}

Let $c \in \mathbb{R}^{C \times T}$ denote the conditioning signal extracted from a wearable window of length $T$, where $C$ is the number of conditioning channels. In our setting, $c$ is constructed from synchronized wearable measurements and derived timing and motion features, as described later in this section. Let $x \in \mathbb{R}^{T}$ denote the corresponding target ECG waveform for the same time window. For notational simplicity, we treat the ECG as a one-channel sequence, although the formulation extends naturally to multi-channel targets.

The goal is to learn a conditional generative mapping from wearable observations to ECG, \(p(x \mid c)\), such that the generated waveform preserves both signal-level fidelity and waveform morphology. Direct deterministic regression seeks a one-shot mapping $f_\theta(c) \rightarrow x$, which can be brittle when the conditioning signals are indirect, noisy, or corrupted by motion. We instead model ECG reconstruction as a continuous-time conditional transport process from a simple reference distribution to the ECG data distribution.

Formally, each training example is a paired sample \((x,c) \sim p_{\mathrm{data}}(x,c)\). We introduce a latent variable $z \sim \mathcal{N}(0,I)$ with the same dimensionality as $x$, and define a continuous family of intermediate states $\{x_t\}_{t \in [0,1]}$ connecting noise to signal. The learning problem is to estimate a time-dependent conditional velocity field $v_\theta(x_t,t,c)$ that transports a sample from the noise distribution toward the target ECG while remaining conditioned on the wearable observations $c$.

\subsection{Conditional rectified-flow formulation}
\label{sec:rf}

We use rectified flow as the generative backbone of BeatFlow-ECG. Rectified flow learns a velocity field along straight conditional paths between a reference sample and a data sample using flow matching \citep{lipman2023flowmatchinggenerativemodeling,liu2022flowstraightfastlearning,esser2024scalingrectifiedflowtransformers,zhang2025trajectoryflowmatchingapplications}. Compared with diffusion-style denoising, sampling is performed by integrating the learned velocity field over an ODE trajectory.

For each paired training example $(x,c)$ and noise sample $z \sim \mathcal{N}(0,I)$, we define the linear interpolation
\begin{equation}
x_t = (1-t)z + tx, \qquad t \in [0,1].
\label{eq:rf_path}
\end{equation}
This path connects pure noise at $t=0$ to the target ECG at $t=1$. Since its time derivative is constant, the network learns to predict the conditional velocity \(v_\theta(x_t,t,c) \approx x-z\).

We train the model using an L1 velocity-matching objective on the rectified-flow target:
\begin{equation}
\mathcal{L}_{\mathrm{FM}}(\theta)
=
\mathbb{E}_{(x,c)\sim p_{\mathrm{data}},\, z\sim \mathcal{N}(0,I),\, t\sim p(t)}
\left[
\left\| v_\theta(x_t, t, c) - (x-z) \right\|_1
\right].
\label{eq:fm_loss}
\end{equation}
The time-sampling distribution $p(t)$ is specified in Appendix~\ref{app:train}.

Intuitively, the model is trained to predict the instantaneous direction that moves the intermediate state $x_t$ toward the target ECG $x$, while conditioning this prediction on the wearable observation $c$.

After training, ECG reconstruction is performed by solving the conditional ordinary differential equation
\begin{equation}
\frac{d x_t}{dt} = v_\theta(x_t, t, c), \qquad x_0 \sim \mathcal{N}(0,I),
\label{eq:rf_ode}
\end{equation}
from $t=0$ to $t=1$. In practice, we use an explicit Euler discretization with $N$ steps:
\begin{equation}
x^{(k+1)} = x^{(k)} + \Delta t \, v_\theta\!\left(x^{(k)}, t_k, c\right),
\qquad
t_k = \frac{k}{N}, \quad \Delta t = \frac{1}{N},
\label{eq:euler_update}
\end{equation}
starting from $x^{(0)} \sim \mathcal{N}(0,I)$. The final state $x^{(N)}$ is taken as the reconstructed ECG.

The conditioning signal $c$ guides the velocity field throughout the transport process. We next describe the architecture used to parameterize $v_\theta(x_t,t,c)$.

\subsection{BeatFlow-ECG architecture}
\label{sec:arch}

\begin{figure}[t]
\begin{center}
\includegraphics[width=\textwidth]{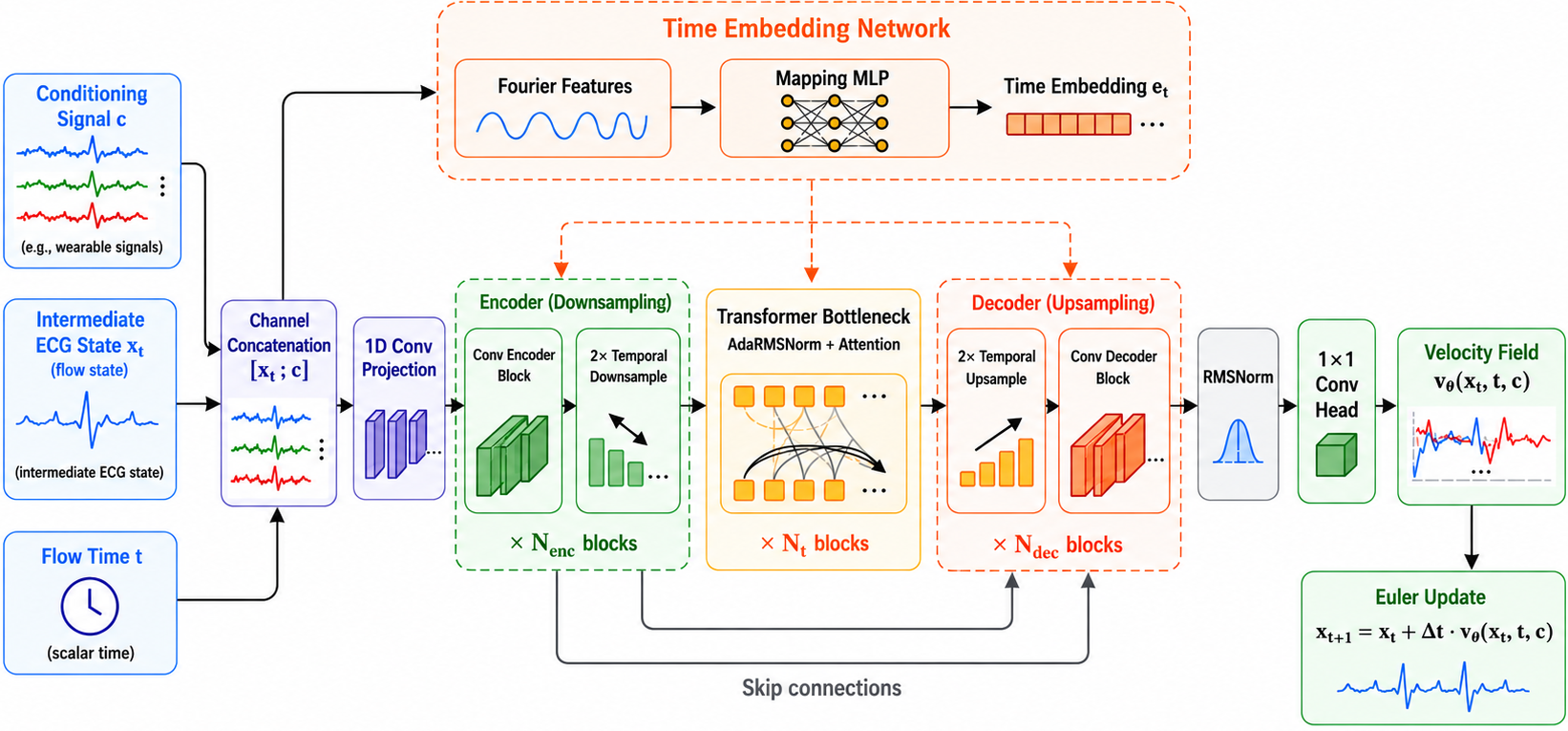}

    \caption{\textbf{Overview of BeatFlow-ECG.} Given the current intermediate state $x_t$, conditioning signal $c$, and flow time $t$, the model predicts a conditional velocity field $v_\theta(x_t,t,c)$ through a convolutional encoder, transformer bottleneck, and temporal decoder.}
    \label{fig:beatflow_overview}
\end{center}
\end{figure}

\begin{figure}[t]
\begin{center}
\includegraphics[width=\textwidth]{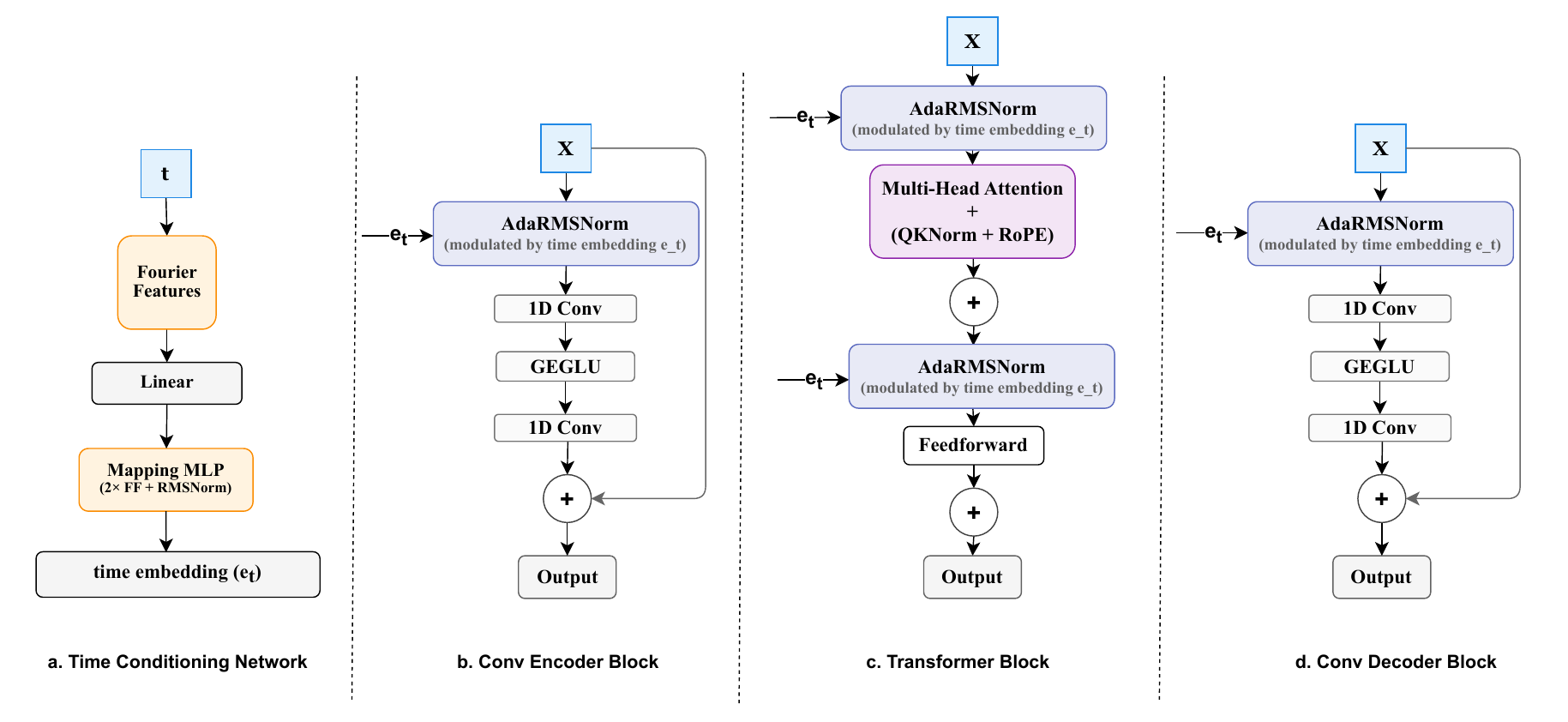}

    \caption{\textbf{Detailed blocks of BeatFlow-ECG.} (a) Time-conditioning network. (b) Convolutional encoder block. (c) Transformer block. (d) Convolutional decoder block. Together, these components parameterize the conditional velocity field used for ECG reconstruction.}
    \label{fig:beatflow_blocks}
\end{center}
\end{figure}

We parameterize $v_\theta(x_t,t,c)$ with a hybrid convolutional-transformer architecture that combines multiscale temporal convolutions with global sequence modeling, similar to recent rectified-flow architectures for signal reconstruction \citep{loraflow}. The model receives the intermediate state $x_t$, wearable conditioning signal $c$, and flow time $t$, where $c$ contains the synchronized wearable measurements and derived timing and motion features. Figures~\ref{fig:beatflow_overview} and~\ref{fig:beatflow_blocks} show the architecture and its main computational blocks; additional details are provided in Appendix~\ref{app:arch}.

\paragraph{Overview.}
BeatFlow-ECG follows an encoder-decoder architecture with a transformer bottleneck. The current state $x_t \in \mathbb{R}^{T}$ is concatenated with the conditioning signal $c \in \mathbb{R}^{C \times T}$ along the channel dimension and projected into the model feature space:
\begin{equation}
h_0 = \mathrm{Conv}_{\mathrm{in}}([x_t; c]).
\end{equation}
This early fusion lets the network learn cross-modal interactions between the current reconstruction state and the wearable observations from the first layer onward.

The fused representation is processed by a temporal encoder that reduces sequence length and extracts local temporal features. Skip connections pass features at finer temporal resolutions to the decoder. A transformer bottleneck models longer-range dependencies across the encoded sequence, including relationships between beats.

\paragraph{Time-conditioned transformer blocks.}
Because the target velocity depends on flow time $t$, the model conditions its intermediate activations on a learned time embedding. We map the scalar time value through Fourier features followed by a small multilayer perceptron:
\begin{equation}
e_t = \mathrm{MLP}(\mathrm{Fourier}(t)).
\end{equation}
The embedding $e_t$ is injected through adaptive RMS normalization (AdaRMSNorm), allowing the network to adapt its behavior across the flow trajectory. The transformer blocks use multi-head self-attention with rotary positional encoding (RoPE) and query-key normalization (QKNorm) for stable temporal modeling, together with GEGLU feedforward blocks for gated feature selection.

\paragraph{Decoder and output head.}
The decoder progressively upsamples the latent representation back to the original temporal resolution and combines it with encoder skip features. A final normalization layer and $1 \times 1$ convolutional head map the decoded representation to the velocity estimate
\begin{equation}
\hat{v} = v_\theta(x_t,t,c).
\end{equation}
The output has the same temporal dimensionality as the target ECG and is used in the Euler update in Equation~\ref{eq:euler_update}.

\subsection{Motion-aware conditioning and training}
\label{sec:motionaware}

Wrist PPG is often corrupted by motion artifacts during unconstrained daily activity, which can distort pulse morphology and weaken the relationship between peripheral pulse measurements and ECG structure \citep{Allen2007-vn,Dias2018-kq,LEMAY2014105,Elgendi2024-iq}. BeatFlow-ECG therefore incorporates motion information through conditioning features, sample weighting, and an easy-to-hard training curriculum. Additional implementation details are provided in Appendix~\ref{app:motionaware}.

\FloatBarrier

\paragraph{Motion-aware conditioning.}
We augment the wearable conditioning representation with IMU-derived descriptors, using acceleration magnitude \(m_{\mathrm{acc}}[n]=\sqrt{a_x[n]^2+a_y[n]^2+a_z[n]^2}\) as a compact measure of instantaneous motion intensity. This provides motion context that can help the model adapt reconstruction under changes in wearable signal quality.

\paragraph{Motion-aware loss weighting.}
To place greater emphasis on higher-motion windows, we compute a window-level score \(s_{\mathrm{motion}}=\mathrm{Std}(m_{\mathrm{acc}})\), normalize it within each mini-batch to \(\tilde{s}_{\mathrm{motion}}\), and set the sample weight to \(w=1+\alpha\tilde{s}_{\mathrm{motion}}\). The weighted flow-matching objective is
\begin{equation}
\mathcal{L}_{\mathrm{motion}}(\theta)
=
\mathbb{E}
\left[
w \, \|v_\theta(x_t,t,c) - (x-z)\|_1
\right].
\label{eq:motion_loss}
\end{equation}

\paragraph{Spectral consistency.}
We additionally use an STFT magnitude loss to encourage consistency in the frequency-domain structure of the reconstructed ECG. For this auxiliary training loss, we define the ECG estimate \(\hat{x}=z+v_\theta(x_t,t,c)\), which equals the target $x$ when the predicted velocity matches the target velocity $x-z$. The full training objective is
\begin{equation}
\mathcal{L}
=
\mathcal{L}_{\mathrm{motion}}
+
\lambda_{\mathrm{STFT}}
\left\|
|\mathrm{STFT}(\hat{x})|-|\mathrm{STFT}(x)|
\right\|_1.
\label{eq:full_loss}
\end{equation}
We use \(\lambda_{\mathrm{STFT}}=0.10\); additional implementation details are provided in Appendix~\ref{app:train}.

\paragraph{Curriculum over motion difficulty.}
We also use an easy-to-hard curriculum in which lower-motion windows are emphasized earlier in training and higher-motion windows are gradually introduced. This helps the model first learn the core physiological mapping before adapting to more corrupted wearable observations.

\subsection{Signal preprocessing and feature construction}
\label{sec:preprocessing}

We preprocess all sensor streams into synchronized fixed-length windows and augment them with timing and motion features. Additional preprocessing details, including the full equations, are provided in Appendix~\ref{app:preproc}.

\paragraph{Filtering, resampling, and normalization.}
Raw PPG and ECG signals contain baseline drift, high-frequency noise, and motion-related perturbations that can obscure physiologically relevant structure \citep{Allen2007-vn,electronics3020282,Li2021-gk,Li_2024}. We apply signal-specific bandpass filtering before cross-modal alignment, using 0.4-8 Hz for PPG and 0.5-40 Hz for ECG. All modalities are then resampled to 125 Hz and z-score normalized independently within each 10-second window.

\paragraph{Alignment and window extraction.}
After preprocessing, all modalities are temporally aligned and segmented into overlapping 10-second windows with a 5-second stride, corresponding to 1250 samples and 625 samples at 125 Hz. Each training example consists of a synchronized wearable conditioning tensor and a one-channel ECG target waveform extracted from the same interval.

\paragraph{Physiological timing and motion features.}
To provide beat-level temporal context, we augment the conditioning signal with a normalized phase feature derived from consecutive PPG peaks \citep{Charlton2022-ac}. This feature represents the relative position of each sample within the pulse cycle. Because wearable PPG is sensitive to motion corruption, we also include IMU-derived motion descriptors, including acceleration magnitude and short-term motion statistics computed over a rolling 0.5-second window. The final conditioning tensor therefore combines synchronized wearable measurements with derived timing and motion features.

\section{Experimental Setup}

We evaluate BeatFlow-ECG on subject-independent ECG reconstruction from wearable signals. Additional implementation and experimental details are provided in Appendices~\ref{app:impl} and~\ref{app:exp}.

\subsection{Datasets and evaluation protocol}
\label{sec:Datasets}

We evaluate on two public wearable datasets, PPG-DaLiA \citep{ppg-dalia_495} and WESAD \citep{10.1145/3242969.3242985}, using a separate LOSO evaluation for each dataset.

\paragraph{PPG-DaLiA.}
PPG-DaLiA contains wrist PPG, wrist inertial measurements, and chest reference signals, including ECG, from 15 subjects during daily-life activities with substantial motion and cross-subject variability. Additional preprocessing and target-construction details are provided in Appendix~\ref{app:preproc}.

\paragraph{WESAD.}
WESAD is a multimodal wearable dataset with synchronized physiological, motion, and ECG signals collected under a different acquisition protocol. We use it as a second benchmark beyond PPG-DaLiA.

We use the preprocessing pipeline in Section~\ref{sec:preprocessing}, with 10-second windows containing PPG, IMU-derived features, and physiological phase information paired with a one-channel ECG target.

We use a leave-one-subject-out (LOSO) protocol on each dataset separately. In each fold, one subject is held out for testing and the remaining subjects are used for training and validation. The held-out subject is not used during model fitting or checkpoint selection, so the evaluation measures generalization to an unseen subject.

\subsection{Baselines and metrics}

We compare BeatFlow-ECG with deterministic, diffusion-based, and adversarial baselines. UNet1D and AttnSeq are convolutional and attention-based direct regressors, respectively; UNet1D-Large and AttnSeq-Large provide capacity-matched variants. Conditional DDPM-1D uses the same conditioning inputs and LOSO protocol as BeatFlow-ECG. CardioGAN \citep{Sarkar_Etemad_2021} uses PPG only, following its original formulation, and is evaluated under the same dataset-specific LOSO protocol on both datasets. Full architecture and training details are provided in Appendix~\ref{app:baseline_details}.

We report L1 error, RMSE, and Pearson correlation $r$ for waveform fidelity, and R-peak F1 with a 100-ms matching tolerance and absolute heart-rate error for beat timing. On PPG-DaLiA, we also report correlations within fixed R-peak-relative windows corresponding approximately to the P-wave, QRS complex, and T-wave regions, together with reconstruction error across IMU-derived motion levels. Additional metric definitions and inference-efficiency results are provided in Appendices~\ref{app:metrics} and~\ref{app:efficiency}.

\subsection{Implementation details}

All models use 125~Hz windows of 1250 samples. In each LOSO fold, 10\% of non-test windows are sampled for validation using seed 42, with the remainder used for training. Because consecutive windows overlap by 5 seconds, training and validation sets can share raw samples within the training subjects, while the held-out test subject remains fully isolated. BeatFlow-ECG uses 120 explicit Euler steps at inference. All baselines use the same dataset-specific LOSO splits and evaluation metrics. Training and baseline details are provided in Appendices~\ref{app:train} and~\ref{app:baseline_details}. Experiments were run on an NVIDIA H100 GPU.

\section{Results}

\begin{table}[H]
\centering
\caption{\textbf{Overall LOSO performance across PPG-DaLiA and WESAD.} Values are mean [95\% CI] across held-out subject folds. Lower is better for L1, RMSE, and HR error; higher is better for Pearson $r$ and R-peak F1.}
\label{tab:overall_comparison}
\resizebox{\columnwidth}{!}{%
\begin{tabular}{llcccccc}
\toprule
\textbf{Dataset} & \textbf{Model} & \textbf{Params} & \textbf{L1} $\downarrow$ & \textbf{RMSE} $\downarrow$ & \textbf{Pearson $r$} $\uparrow$ & \textbf{R-peak F1} $\uparrow$ & \textbf{HR error (bpm)} $\downarrow$ \\
\midrule
\multirow{7}{*}{PPG-DaLiA} 
& AttnSeq & 8.5M & 0.285 [0.258, 0.312] & 0.412 [0.375, 0.449] & 0.815 [0.791, 0.839] & 0.742 [0.717, 0.767] & 10.45 [9.47, 11.43] \\
& AttnSeq-Large & 12.3M & 0.258 [0.231, 0.285] & 0.381 [0.342, 0.420] & 0.835 [0.810, 0.860] & 0.768 [0.740, 0.796] & 9.31 [8.40, 10.22] \\
& UNet1D & 4.2M & 0.124 [0.114, 0.134] & 0.185 [0.172, 0.198] & 0.931 [0.918, 0.944] & 0.885 [0.873, 0.897] & 6.72 [6.14, 7.30] \\
& UNet1D-Large & 13.2M & 0.138 [0.086, 0.190] & 0.210 [0.138, 0.282] & 0.925 [0.872, 0.978] & 0.875 [0.832, 0.918] & 6.95 [5.68, 8.22] \\
& Conditional DDPM-1D & 12.6M & 0.085 [0.067, 0.103] & 0.132 [0.111, 0.153] & 0.962 [0.946, 0.979] & 0.905 [0.882, 0.928] & 6.12 [5.42, 6.82] \\
& CardioGAN & 28.24M & 0.377 [0.332, 0.422] & 0.465 [0.413, 0.517] & 0.810 [0.772, 0.848] & 0.755 [0.720, 0.790] & 10.20 [8.95, 11.45] \\
& \textbf{BeatFlow-ECG} & \textbf{12.5M} & \textbf{0.062 [0.051, 0.072]} & \textbf{0.102 [0.089, 0.116]} & \textbf{0.983 [0.971, 0.994]} & \textbf{0.946 [0.933, 0.958]} & \textbf{4.89 [4.41, 5.37]} \\
\midrule
\multirow{7}{*}{WESAD}
& AttnSeq & 8.5M & 0.240 [0.215, 0.265] & 0.365 [0.330, 0.400] & 0.840 [0.815, 0.865] & 0.780 [0.750, 0.810] & 9.15 [8.20, 10.10] \\
& AttnSeq-Large & 12.3M & 0.192 [0.171, 0.213] & 0.308 [0.272, 0.344] & 0.884 [0.860, 0.908] & 0.828 [0.798, 0.858] & 7.38 [6.52, 8.24] \\
& UNet1D & 4.2M & 0.105 [0.095, 0.115] & 0.160 [0.145, 0.175] & 0.945 [0.930, 0.960] & 0.902 [0.890, 0.914] & 5.85 [5.30, 6.40] \\
& UNet1D-Large & 13.2M & 0.128 [0.081, 0.175] & 0.194 [0.130, 0.258] & 0.921 [0.856, 0.986] & 0.876 [0.828, 0.924] & 6.42 [5.26, 7.58] \\
& Conditional DDPM-1D & 12.6M & 0.074 [0.061, 0.088] & 0.113 [0.098, 0.129] & 0.971 [0.958, 0.984] & 0.918 [0.899, 0.937] & 5.25 [4.46, 6.04] \\
& CardioGAN & 28.24M & 0.335 [0.295, 0.375] & 0.425 [0.380, 0.470] & 0.825 [0.795, 0.855] & 0.780 [0.745, 0.815] & 9.85 [8.70, 11.00] \\
& \textbf{BeatFlow-ECG} & \textbf{12.5M} & \textbf{0.055 [0.046, 0.064]} & \textbf{0.091 [0.080, 0.102]} & \textbf{0.986 [0.978, 0.995]} & \textbf{0.955 [0.944, 0.965]} & \textbf{4.12 [3.74, 4.50]} \\
\bottomrule
\end{tabular}%
}
\end{table}

Table~\ref{tab:overall_comparison} shows that BeatFlow-ECG achieves the best results across all reported metrics. On PPG-DaLiA, compared with Conditional DDPM-1D, it reduces L1 error from 0.085 to 0.062, increases Pearson correlation from 0.962 to 0.983, and improves R-peak F1 from 0.905 to 0.946. On WESAD, it reduces L1 error from 0.074 to 0.055 and improves R-peak F1 from 0.918 to 0.955. The capacity-matched UNet1D-Large and AttnSeq-Large baselines remain weaker, indicating that parameter count alone does not explain the performance gap.

Paired Wilcoxon signed-rank tests across the 15 held-out PPG-DaLiA subjects show lower L1 and HR error for BeatFlow-ECG than for Conditional DDPM-1D and UNet1D ($p < 0.01$ for each comparison).

\subsection{Morphology-sensitive analysis}

Whole-window correlation can be dominated by the QRS complex and may not fully reflect lower-amplitude ECG morphology. We therefore evaluate Pearson correlation over fixed R-peak-relative windows on PPG-DaLiA that correspond approximately to the P-wave, QRS complex, and T-wave regions. Appendix~\ref{app:segment_analysis} provides the window boundaries.

Table~\ref{tab:morphology_main} shows that BeatFlow-ECG achieves the highest correlation in all three waveform regions, including the lower-amplitude P-wave and T-wave regions. These results are limited to the healthy cohorts used in this study and do not establish generalization to pathological ECG morphology.

\begin{table}[t]
\centering
\caption{\textbf{Morphology-sensitive evaluation on PPG-DaLiA.} Pearson correlation is computed over fixed R-peak-relative windows corresponding approximately to the P-wave, QRS complex, and T-wave regions. Higher is better.}
\label{tab:morphology_main}
\begin{tabular}{lccc}
\toprule
\textbf{Model} &
\textbf{P-wave $r$} $\uparrow$ &
\textbf{QRS $r$} $\uparrow$ &
\textbf{T-wave $r$} $\uparrow$ \\
\midrule
UNet1D & 0.742 & 0.951 & 0.865 \\
UNet1D-Large & 0.712 & 0.935 & 0.838 \\
Conditional DDPM-1D & 0.825 & 0.978 & 0.908 \\
\textbf{BeatFlow-ECG} & \textbf{0.882} & \textbf{0.992} & \textbf{0.945} \\
\bottomrule
\end{tabular}
\end{table}

\begin{figure}[t]
    \centering
    \begin{subfigure}[t]{0.32\textwidth}
        \centering
        \includegraphics[width=\textwidth]{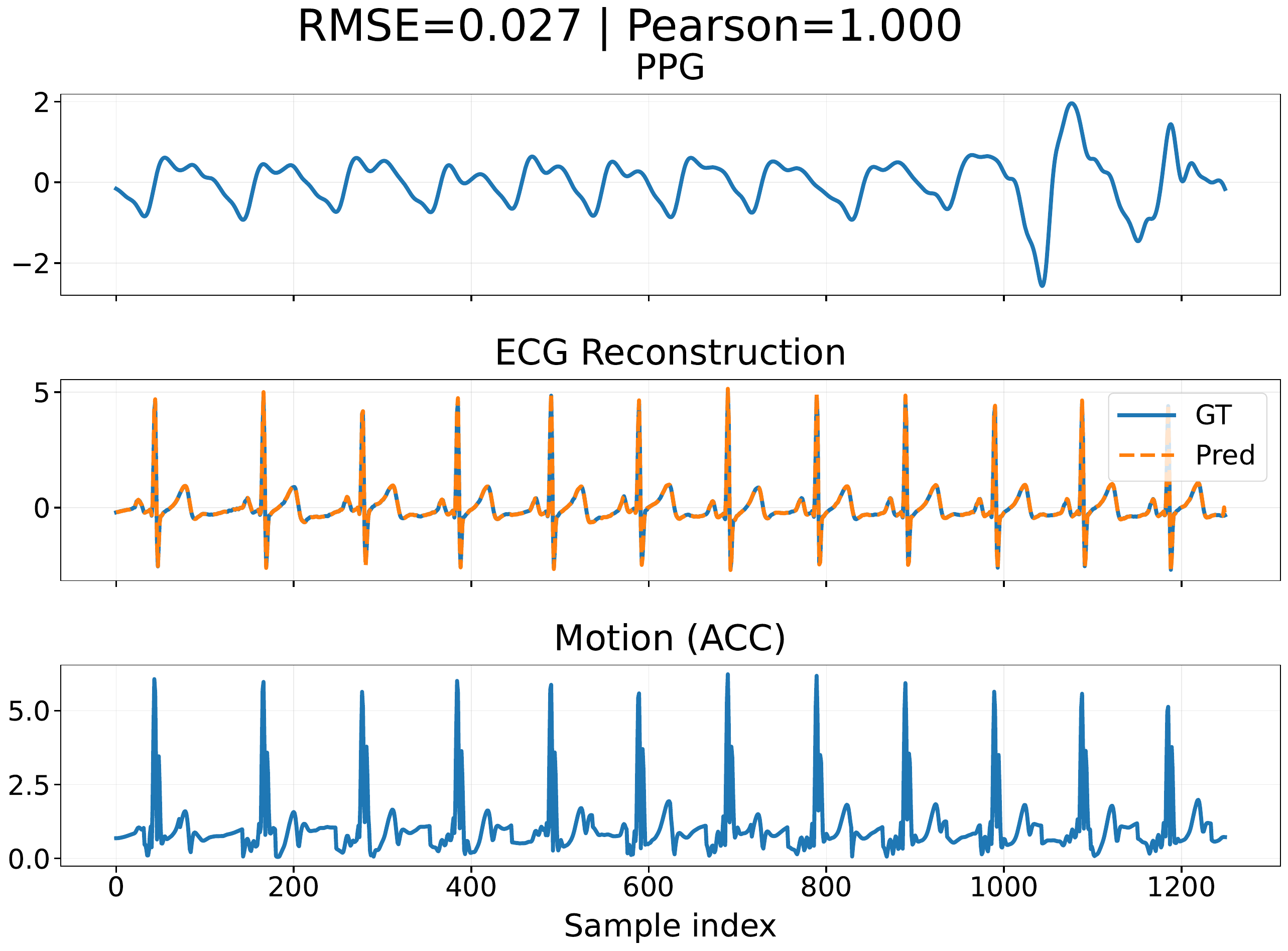}
        \caption{Low-error example.}
        \label{fig:qual_best}
    \end{subfigure}
    \hfill
    \begin{subfigure}[t]{0.32\textwidth}
        \centering
        \includegraphics[width=\textwidth]{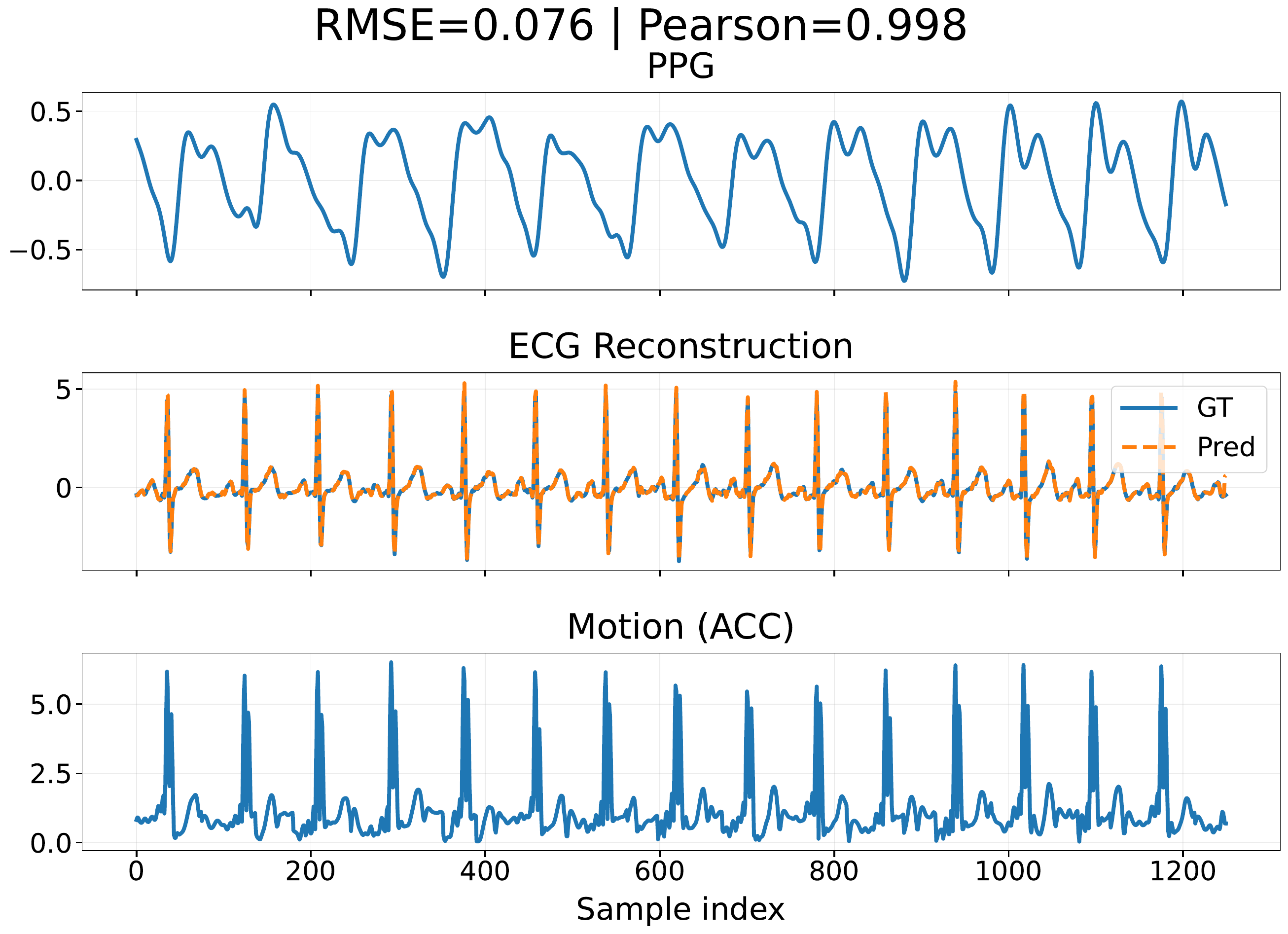}
        \caption{Moderate-error example.}
        \label{fig:qual_mid}
    \end{subfigure}
    \hfill
    \begin{subfigure}[t]{0.32\textwidth}
        \centering
        \includegraphics[width=\textwidth]{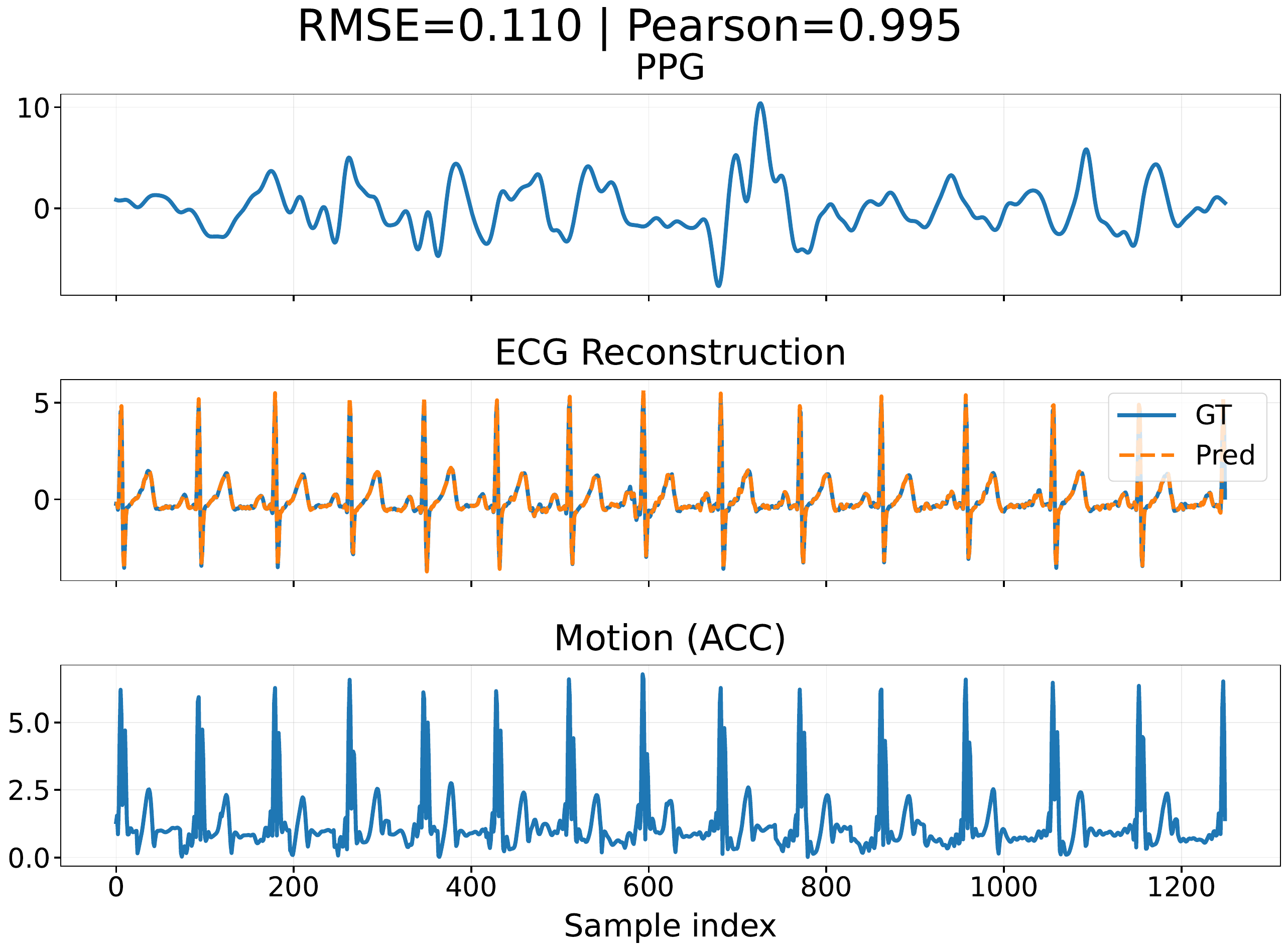}
        \caption{Higher-error example.}
        \label{fig:qual_hard}
    \end{subfigure}

    \caption{\textbf{Qualitative ECG reconstruction examples.}
    Each panel shows wearable PPG, reconstructed ECG, reference ECG, and motion signal across windows with varying reconstruction error.}
    \label{fig:qualitative_examples}
\end{figure}

\subsection{Subject-wise analysis}

BeatFlow-ECG performs strongly across most held-out subjects. S6 and S10 are among the more difficult PPG-DaLiA folds; for S6, the dataset authors also report a recording hardware issue that reduced the valid recording duration \citep{ppg-dalia_495}. We retain both subjects in the evaluation. Detailed subject-wise results are provided in Appendix~\ref{app:quant_subjects}.

\subsection{Qualitative reconstruction analysis}

Figure~\ref{fig:qualitative_examples} shows example reconstructions across windows with varying reconstruction error. BeatFlow-ECG preserves the dominant beat structure, timing, and local waveform morphology. Additional peak-alignment examples, reconstruction trajectories, and baseline comparisons are provided in Appendices~\ref{app:qual_peaks}, \ref{app:qual_traj}, and~\ref{app:quant_baselines}.

\subsection{Motion robustness analysis}

Because PPG-DaLiA contains substantial daily-life motion, we further evaluate reconstruction across low-, medium-, and high-motion subsets. No test windows are excluded based on ECG quality, motion level, or reconstruction difficulty; all windows satisfying the fixed preprocessing and alignment requirements are retained. For each held-out test window, we compute the same IMU-derived motion score and divide the test windows into low-, medium-, and high-motion groups using the 33rd and 67th percentiles of the motion-score distribution. The resulting grouping is fixed and applied identically to all models.

\begin{table}[t]
\centering
\caption{\textbf{Motion-stratified L1 error on PPG-DaLiA.} Lower is better.}
\label{tab:motion_stratified}
\begin{tabular}{lccc}
\toprule
\textbf{Motion level} &
\textbf{UNet1D-Large} &
\textbf{Conditional DDPM-1D} &
\textbf{BeatFlow-ECG} \\
\midrule
Low    & 0.070 & 0.045 & \textbf{0.032} \\
Medium & 0.127 & 0.075 & \textbf{0.054} \\
High   & 0.217 & 0.135 & \textbf{0.100} \\
\bottomrule
\end{tabular}
\end{table}

Table~\ref{tab:motion_stratified} shows that reconstruction error increases with motion for all models, while BeatFlow-ECG remains lowest across all three motion levels. The IMU-derived score measures motion exposure only and is not interpreted as ECG SNR, since PPG-DaLiA does not provide a paired artifact-free ECG reference for each segment. Additional motion visualizations are provided in Appendix~\ref{app:motion_analysis}.

\subsection{Ablation Study}

We evaluate the contribution of the main design components of BeatFlow-ECG on PPG-DaLiA. Each variant removes the indicated component while keeping the remaining training setup fixed.

\begin{table}[t]
\centering
\caption{\textbf{Ablation study on PPG-DaLiA.} Lower is better for L1, RMSE, and HR error; higher is better for Pearson $r$ and R-peak F1.}
\label{tab:ablation_main}
\resizebox{\columnwidth}{!}{%
\begin{tabular}{lccccc}
\toprule
\textbf{Configuration} &
\textbf{L1} $\downarrow$ &
\textbf{RMSE} $\downarrow$ &
\textbf{Pearson $r$} $\uparrow$ &
\textbf{R-peak F1} $\uparrow$ &
\textbf{HR error (bpm)} $\downarrow$ \\
\midrule
w/o Motion Conditioning & 0.078 & 0.125 & 0.965 & 0.920 & 5.60 \\
w/o Curriculum & 0.071 & 0.118 & 0.974 & 0.932 & 5.35 \\
w/o STFT Loss & 0.069 & 0.116 & 0.975 & 0.926 & 5.21 \\
w/o Motion Weighting + STFT Loss & 0.084 & 0.132 & 0.960 & 0.912 & 5.85 \\
\midrule
\textbf{BeatFlow-ECG (Full)} &
\textbf{0.062} &
\textbf{0.102} &
\textbf{0.983} &
\textbf{0.946} &
\textbf{4.89} \\
\bottomrule
\end{tabular}%
}
\end{table}

Table~\ref{tab:ablation_main} shows that all evaluated ablations degrade performance. Removing both motion weighting and the STFT loss gives the largest degradation, increasing L1 from 0.062 to 0.084 and HR error from 4.89 to 5.85 bpm. Motion conditioning, the curriculum, and the STFT objective also each improve performance.

\section{Discussion}

BeatFlow-ECG remains stronger than the deterministic baselines and Conditional DDPM-1D, while the capacity-matched variants indicate that model size alone does not explain the performance gap. On PPG-DaLiA, BeatFlow-ECG also shows higher regional waveform correlations and lower reconstruction error across motion levels. Although slower than the single-pass deterministic models, it requires substantially fewer function evaluations than the evaluated DDPM configuration (Appendix~\ref{app:efficiency}).

\section{Limitations}

Our evaluation is limited to PPG-DaLiA and WESAD and does not establish generalization across different devices, populations, sensor placements, or clinical conditions. Both datasets mainly contain healthy subjects, so the morphology results do not demonstrate reconstruction of pathological ECG patterns such as atrial fibrillation, conduction abnormalities, or abnormal T-wave morphology. The current model also reconstructs a single-channel ECG rather than a diagnostic 12-lead ECG. Future work should evaluate larger and more clinically diverse cohorts, investigate uncertainty estimation, and conduct prospective clinical validation.

\section{Conclusion}

We introduced BeatFlow-ECG, a conditional rectified-flow framework for ECG reconstruction from indirect wearable signals. Under LOSO on PPG-DaLiA and WESAD, it achieves the best overall waveform and beat-timing results among the evaluated baselines, with higher regional waveform correlations and lower reconstruction error across motion levels on PPG-DaLiA.

\section*{Ethics Statement}

BeatFlow-ECG reconstructs ECG rather than directly measuring it, and reconstruction errors could lead to false reassurance or false alarms if used clinically. The results in this study should therefore not be interpreted as establishing diagnostic validity. Prospective clinical validation would be required before clinical use.

\section*{Reproducibility Statement}

The main paper describes the model formulation, preprocessing pipeline, LOSO evaluation protocol, baselines, and evaluation metrics. The appendix provides additional architecture and training details, preprocessing equations, baseline configurations, confidence intervals, subject-wise results, and efficiency analyses. All experiments use the publicly available PPG-DaLiA and WESAD datasets. 

\section*{AI Use Statement}
We used LLMs to polish the manuscript and correct grammatical errors. All scientific content was developed and verified by the authors.

\bibliography{iclr2027_conference}
\bibliographystyle{iclr2027_conference}

\appendix
\section*{Appendix}
\section{Additional implementation and training details}
\label{app:impl}

\subsection{Detailed architecture configuration}
\label{app:arch}

BeatFlow-ECG uses the conditional rectified-flow architecture described in Section~\ref{sec:arch}. In the configuration used for the LOSO experiments, the model reconstructs a one-channel ECG waveform from synchronized wearable signals over windows of 1250 samples. The conditioning representation follows Section~\ref{sec:preprocessing} and includes PPG, IMU-derived channels, and a physiological phase feature, with motion-aware conditioning and training as described in Section~\ref{sec:motionaware}.

The backbone uses 8 transformer layers with hidden dimension 320 and feedforward dimension 1280, with dropout set to 0.0 and skip connections enabled. The conditioning signal and current state are fused before the convolutional encoder and transformer bottleneck, with flow-time conditioning applied through AdaRMSNorm as described in the main text.

The implementation includes temporal-alignment safeguards for small length mismatches introduced by downsampling and upsampling. Intermediate tensors are kept in a consistent channel-first layout, aligned in temporal length when necessary, and skip connections are applied only after temporal matching.

\subsection{Training objective and optimization details}
\label{app:train}

BeatFlow-ECG follows the conditional rectified-flow formulation in Section~\ref{sec:rf}. For each training sample, the flow time $t$ is drawn from a $\mathrm{Beta}(0.5,0.5)$ distribution and clamped to a minimum of $10^{-3}$ to avoid boundary cases. Let $x$ denote the target ECG waveform and $z \sim \mathcal{N}(0,I)$ Gaussian noise of the same shape. The intermediate state is
\begin{equation}
x_t = (1-t)z + tx,
\end{equation}
with target velocity
\begin{equation}
v = x - z.
\end{equation}
The model is trained to predict the conditional velocity field from $(x_t,t,c)$, where $c$ denotes the wearable conditioning signal.

The full model combines a motion-weighted L1 velocity-matching loss with an STFT magnitude loss to encourage spectral consistency in the reconstructed waveform. Because the network predicts a velocity field rather than the final ECG directly, we compute the STFT term on the implied ECG estimate
\begin{equation}
\hat{x} = z + v_\theta(x_t,t,c),
\end{equation}
which equals the target $x$ when the predicted velocity matches the rectified-flow target $x-z$. The spectral term is
\begin{equation}
\mathcal{L}_{\mathrm{STFT}}
=
\left\|
|\mathrm{STFT}(\hat{x})| -
|\mathrm{STFT}(x)|
\right\|_1.
\end{equation}
The final training objective for the full model is
\begin{equation}
\mathcal{L}
=
\mathcal{L}_{\mathrm{motion}}
+
\lambda_{\mathrm{STFT}}\mathcal{L}_{\mathrm{STFT}},
\end{equation}
with $\lambda_{\mathrm{STFT}}=0.10$ and $n_{\mathrm{fft}}=256$. For the \emph{w/o Motion Weighting + STFT Loss} ablation, $\mathcal{L}_{\mathrm{motion}}$ is replaced by the unweighted velocity-matching loss $\mathcal{L}_{\mathrm{FM}}$, and the STFT term is removed. Gradient norms are clipped to 1.0 for stability.

We optimize BeatFlow-ECG using AdamW with learning rate $2\times10^{-4}$, batch size 128, and 80 training epochs. A cosine annealing learning-rate schedule is used over the full training horizon. Within each LOSO fold, checkpoints are selected using validation windows from non-test subjects. Because consecutive windows overlap within recordings, training and validation windows can share raw samples. Validation scores are used for checkpoint selection, while all reported test metrics are computed on the held-out subject, which is excluded from both training and validation.

The standard UNet1D and AttnSeq baselines are trained as direct regressors using the same LOSO splits and preprocessing pipeline. Both use AdamW with learning rate $10^{-3}$, batch size 64, 30 training epochs, and L1 reconstruction loss. Training settings for the larger variants are provided in Appendix~\ref{app:baseline_details}. Each model is trained for the full epoch schedule, and the checkpoint with the best validation performance is selected within each LOSO fold using only non-test subjects. In our runs, both standard models reached their best validation performance before epoch 30 and had plateaued by the end of training.

At inference time, BeatFlow-ECG starts from Gaussian noise and reconstructs the ECG waveform by integrating the learned velocity field with an explicit Euler solver for 120 steps.

\subsection{Motion-aware training details}
\label{app:motionaware}

This section provides additional details on the motion-aware conditioning, loss weighting, and curriculum described in Section~\ref{sec:motionaware}.

Motion information is incorporated into the conditioning pathway using IMU-derived descriptors. Given preprocessed accelerometer channels $(a_x[n],a_y[n],a_z[n])$, we compute the acceleration magnitude
\begin{equation}
m_{\mathrm{acc}}[n] =
\sqrt{a_x[n]^2 + a_y[n]^2 + a_z[n]^2}.
\end{equation}

Second, the training objective is weighted according to motion difficulty. For each window, we define a scalar motion score from the standard deviation of the acceleration magnitude:
\begin{equation}
s_{\mathrm{motion}} = \mathrm{Std}\!\left(m_{\mathrm{acc}}\right).
\end{equation}
Within each mini-batch, this score is normalized as
\begin{equation}
\tilde{s}_{\mathrm{motion}} =
\frac{s_{\mathrm{motion}} - s_{\min}}
{s_{\max} - s_{\min} + \varepsilon},
\end{equation}
where $s_{\min}$ and $s_{\max}$ denote the minimum and maximum motion scores within the mini-batch. The normalized score is then converted into a per-sample training weight
\begin{equation}
w = 1 + \alpha \tilde{s}_{\mathrm{motion}},
\end{equation}
where $\alpha$ controls the emphasis placed on high-motion samples; we use $\alpha=1.0$ in all reported experiments. The resulting objective is the weighted flow-matching loss
\begin{equation}
\mathcal{L}_{\mathrm{motion}}(\theta)
=
\mathbb{E}
\left[
w \, \|v_\theta(x_t,t,c) - (x-z)\|_1
\right].
\end{equation}
Higher-motion windows therefore receive larger training weights. We use mini-batch normalization rather than global dataset statistics to avoid an additional dataset-level calibration step and keep the weighting adaptive across folds.

Training also follows an easy-to-hard curriculum based on the motion score $s_{\mathrm{motion}}$. We rank training windows by motion score and use a linear pacing schedule over epochs: early epochs sample primarily from the lower-motion portion of the training set, and the accessible fraction is gradually increased until all training windows are used.

\section{Additional experimental details}
\label{app:exp}

\subsection{Dataset preprocessing details}
\label{app:preproc}

This section describes the preprocessing pipeline used to construct paired wearable-conditioning and ECG-target windows.

\paragraph{Bandpass filtering.}
We apply signal-specific bandpass filtering before cross-modal alignment to reduce baseline drift, high-frequency noise, and motion-related perturbations \citep{Allen2007-vn,electronics3020282,Li2021-gk,Li_2024}. Let $x(t)$ denote a raw signal sampled at rate $f_{\mathrm{in}}$. We define the filtered signal as
\begin{equation}
x_{\mathrm{bp}}(t) = \mathrm{BP}\!\left(x(t); f_\ell, f_h\right),
\end{equation}
where $\mathrm{BP}(\cdot)$ denotes a bandpass operator and $(f_\ell,f_h)$ are the low- and high-cutoff frequencies. For PPG, we use $(f_\ell,f_h)=(0.4,8)$ Hz to preserve pulse dynamics while attenuating slow drift and high-frequency noise. For ECG, we use $(f_\ell,f_h)=(0.5,40)$ Hz to preserve clinically relevant morphology, including sharp QRS energy, while removing baseline wander and higher-frequency artifacts.

\paragraph{Resampling to a common rate.}
The raw sensing streams originate from different devices and may operate at different sampling frequencies. All modalities are resampled to a common frequency of $f_s = 125$ Hz:
\begin{equation}
x_{125}[n] = \mathcal{R}\!\left(x_{\mathrm{bp}}; f_{\mathrm{in}} \rightarrow f_s\right),
\end{equation}
where $\mathcal{R}(\cdot)$ denotes rational resampling. This gives the target ECG and conditioning channels a common sampling interval.

\paragraph{Per-channel normalization.}
After resampling, each channel is z-score normalized independently within each 10-second window using statistics computed from that window. For a signal $x[n]$ of length $N$, we compute
\begin{equation}
\tilde{x}[n] = \frac{x[n] - \mu_x}{\sigma_x + \varepsilon},
\qquad
\mu_x = \frac{1}{N}\sum_{n=1}^{N} x[n],
\qquad
\sigma_x = \sqrt{\frac{1}{N}\sum_{n=1}^{N}(x[n]-\mu_x)^2},
\end{equation}
where $\varepsilon > 0$ is a small numerical constant. This normalization is applied independently to the PPG channel, the selected ECG target, and each IMU-derived channel.

\begin{figure}[t]
    \centering
    \begin{subfigure}[t]{0.32\textwidth}
        \centering
        \includegraphics[width=\textwidth]{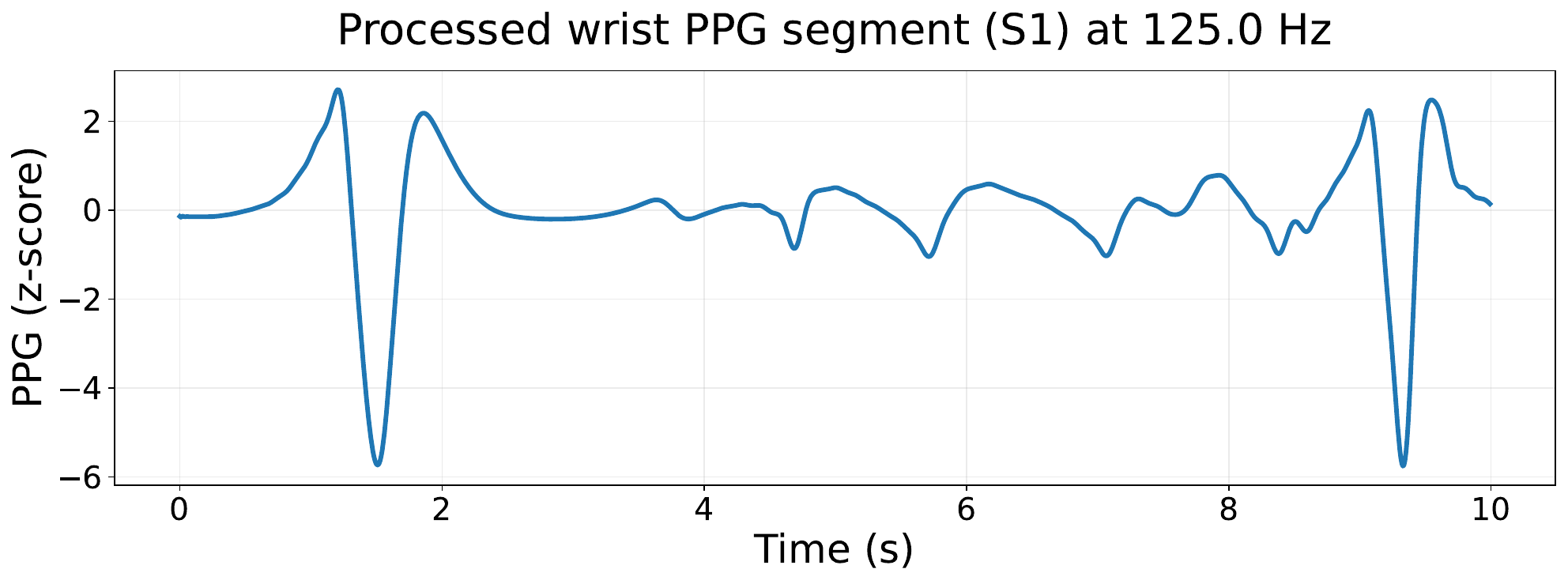}
        \caption{Processed wrist PPG.}
        \label{fig:dalia_ppg}
    \end{subfigure}
    \hfill
    \begin{subfigure}[t]{0.32\textwidth}
        \centering
        \includegraphics[width=\textwidth]{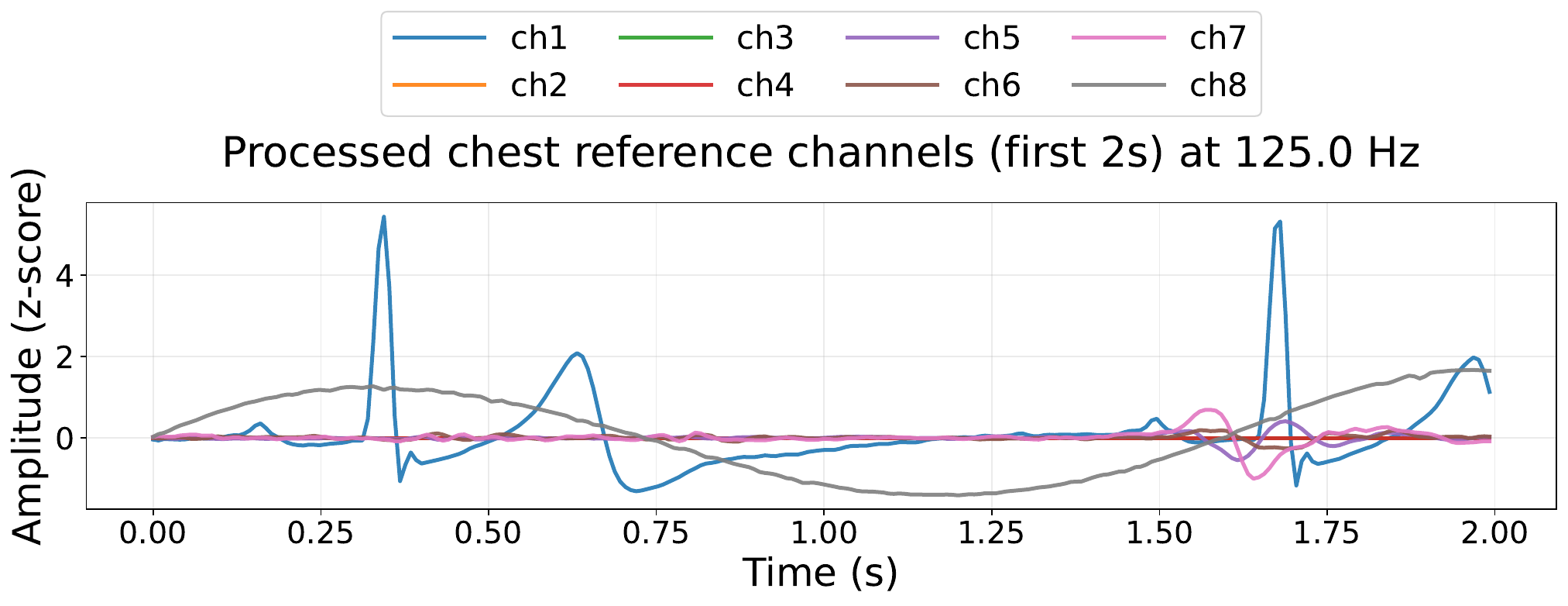}
        \caption{RespiBAN reference channels.}
        \label{fig:dalia_respiban}
    \end{subfigure}
    \hfill
    \begin{subfigure}[t]{0.32\textwidth}
        \centering
        \includegraphics[width=\textwidth]{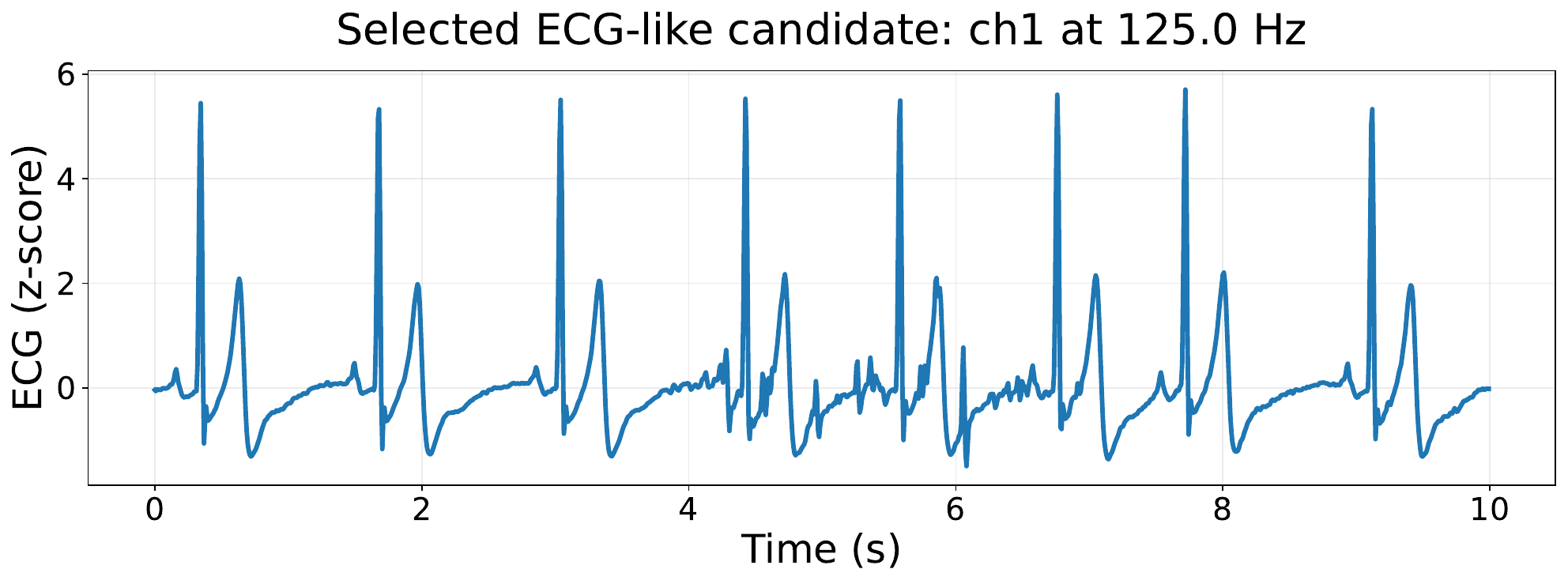}
        \caption{Selected ECG target.}
        \label{fig:dalia_target}
    \end{subfigure}

    \caption{\textbf{Wearable ECG reconstruction setting on PPG-DaLiA.}
    (a) Processed wrist PPG segment. (b) Processed chest reference channels from the RespiBAN device. (c) Selected one-channel ECG target used for supervision.}
    \label{fig:dalia_overview}
\end{figure}

\paragraph{ECG target selection.}
For PPG-DaLiA, we use the ECG channel provided by the RespiBAN chest device as the reconstruction target \citep{ppg-dalia_495}. In the synchronized dataset, this corresponds to \texttt{signal/chest/ECG} and is used for all windows. Respiration and other non-ECG chest signals are not used as reconstruction targets. After filtering and resampling, the ECG channel is segmented into the same 10-second windows as the wearable conditioning signals. Each ECG window is then z-score normalized using its own mean and standard deviation. Figure~\ref{fig:dalia_overview} shows the processed wrist PPG, RespiBAN reference signals, and the ECG target used for reconstruction.

\paragraph{Common valid length.}
After resampling, the modalities may have slightly different valid durations. We truncate them to a common length before window extraction. If $L_{\mathrm{PPG}}$, $L_{\mathrm{ECG}}$, and $L_{\mathrm{IMU}}$ denote the available lengths after preprocessing, we define
\begin{equation}
L = \min(L_{\mathrm{PPG}}, L_{\mathrm{ECG}}, L_{\mathrm{IMU}}),
\end{equation}
and retain the first $L$ samples from each modality.

\paragraph{Window extraction.}
The aligned signals are converted into supervised training examples using overlapping sliding windows. We use a window duration of $T_w = 10$ s and stride $S_w = 5$ s, corresponding to
\begin{equation}
N_T = T_w \cdot f_s = 1250,
\qquad
N_S = S_w \cdot f_s = 625
\end{equation}
samples, respectively. Let $\tilde{c}$ denote the wearable conditioning tensor in channel-first format and $y$ the target ECG waveform. For window index $m$, starting at $s_m = mN_S$, we define
\begin{equation}
c_m = \tilde{c}[:,\,s_m : s_m + N_T], \qquad
y_m = y[s_m : s_m + N_T],
\end{equation}
for all valid $m$ such that $s_m + N_T \leq L$. Each training example therefore contains a synchronized wearable conditioning tensor and a target ECG waveform covering the same 10-second interval.

\paragraph{Physiological phase feature.}
To provide beat-level temporal context, we augment the conditioning signal with a phase feature derived from PPG peaks. Let $p_i$ and $p_{i+1}$ denote consecutive detected PPG peak indices \citep{Charlton2022-ac}. For each sample index $n \in [p_i, p_{i+1})$, we define the normalized phase
\begin{equation}
\phi[n] = \frac{n - p_i}{p_{i+1} - p_i},
\qquad
\phi[n] \in [0,1].
\label{eq:phase_feature}
\end{equation}
The feature resets at each pulse cycle and encodes the relative position within the pulse cycle. In implementation, PPG peaks are detected using a local-maximum detector with a minimum inter-peak distance chosen to exclude physiologically implausible heart rates. If fewer than two peaks are detected in a window, the phase channel is set to zero. Samples before the first detected peak are assigned phase 0, and samples after the last detected peak are assigned the final available phase value.

\paragraph{Motion-related features.}
Because wearable PPG is highly sensitive to motion corruption, we include explicit motion descriptors derived from the wrist IMU. Let $(a_x[n],a_y[n],a_z[n])$ denote the accelerometer channels after preprocessing. We compute the acceleration magnitude as
\begin{equation}
m_{\mathrm{acc}}[n] = \sqrt{a_x[n]^2 + a_y[n]^2 + a_z[n]^2}.
\end{equation}
We further compute short-term motion statistics over a rolling 0.5 s window. Specifically, the conditioning tensor includes the rolling standard deviation of the acceleration magnitude over this window to capture local motion variability.

\paragraph{Final conditioning tensor.}
After preprocessing and feature construction, each training window consists of synchronized wearable measurements and derived timing and motion features paired with a target ECG waveform $y_m$.

\subsection{Evaluation metrics}
\label{app:metrics}

We report waveform-level and beat-timing evaluation metrics.

\paragraph{L1 and RMSE.}
Signal-level reconstruction accuracy is measured using L1 error and root mean squared error (RMSE) between the reconstructed and target ECG waveforms over each test window.

\paragraph{Pearson correlation.}
We report the Pearson correlation coefficient between the reconstructed and target ECG to measure waveform similarity independent of scale.

\paragraph{R-peak F1.}
We detect R-peaks in the reconstructed and target ECG and compute precision, recall, and F1 score using a temporal matching tolerance of 100 ms.

\paragraph{Heart-rate error.}
We compute heart rate from the detected peaks and report the absolute error in beats per minute (bpm) between the reconstructed and target signals.

\paragraph{Motion-aware analysis.}
For the motion-stratified analysis, we compute a motion score from the wearable inertial channels and evaluate reconstruction quality across motion conditions.
\begin{figure}[t]
    \centering
    \begin{subfigure}[t]{0.48\textwidth}
        \centering
        \includegraphics[width=\textwidth]{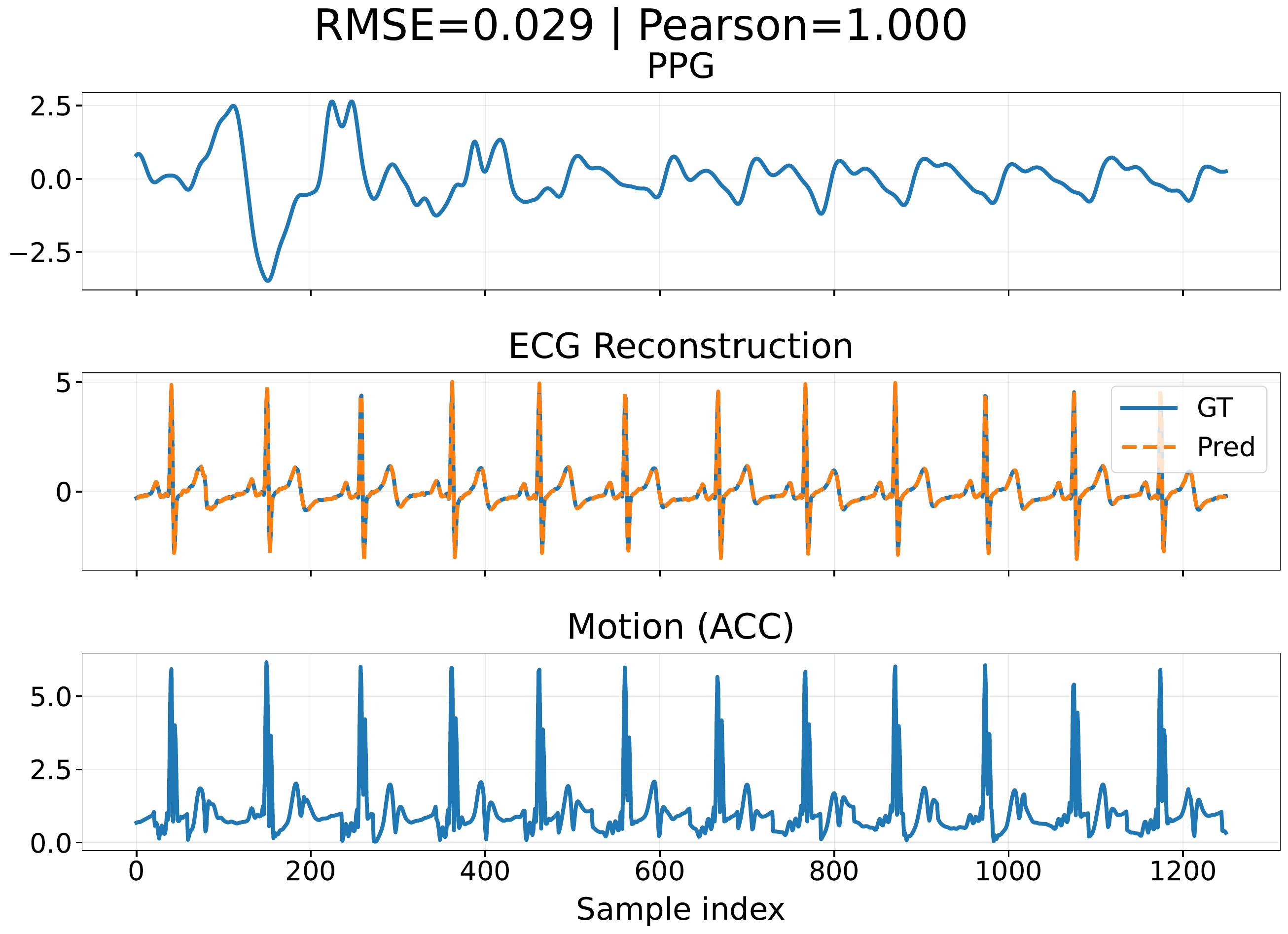}
        \caption{Additional qualitative example 1.}
        \label{fig:app_best_full_27299}
    \end{subfigure}
    \hfill
    \begin{subfigure}[t]{0.48\textwidth}
        \centering
        \includegraphics[width=\textwidth]{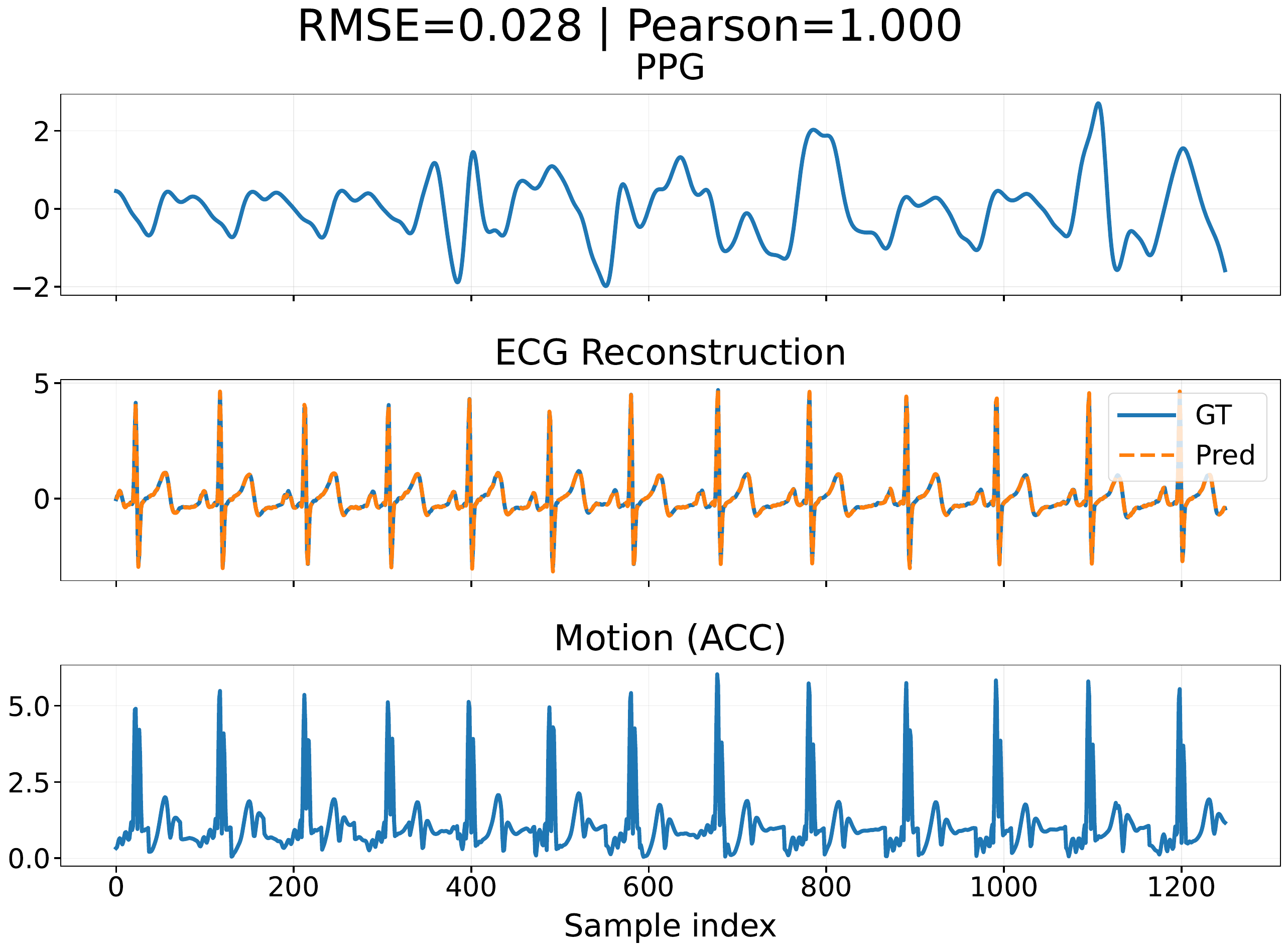}
        \caption{Additional qualitative example 2.}
        \label{fig:app_best_full_27346}
    \end{subfigure}
        \caption{\textbf{Additional BeatFlow-ECG reconstruction examples.}
        Each panel shows the conditioning PPG, reconstructed ECG, ground-truth ECG, and motion signal.}
    \label{fig:app_additional_recons}
\end{figure}

\subsection{Compute resources}
\label{app:compute}

All models were trained and evaluated on an NVIDIA H100 GPU. For each LOSO fold, the experimental pipeline saves checkpoints, predictions, summary metrics, and qualitative outputs.

\subsection{Confidence intervals for LOSO results}
\label{app:ci}

We compute 95\% confidence intervals across the 15 held-out subject folds for each reported model and metric. For an evaluation metric, let $x_1,\dots,x_{15}$ denote the values from the 15 LOSO test folds. We report the mean
\begin{equation}
\bar{x} = \frac{1}{15}\sum_{i=1}^{15} x_i
\end{equation}
together with a 95\% confidence interval of the form
\begin{equation}
\bar{x} \pm t_{0.975,14}\frac{s}{\sqrt{15}},
\end{equation}
where $s$ is the sample standard deviation across folds and $t_{0.975,14}$ is the $97.5$th percentile of the Student-$t$ distribution with 14 degrees of freedom.

These intervals summarize variability across held-out subjects and therefore reflect subject-to-subject variation rather than repeated random-seed variation.

\section{Additional qualitative results}
\label{app:qual}

\subsection{Additional reconstruction examples}
\label{app:qual_examples}

Figure~\ref{fig:app_additional_recons} shows additional BeatFlow-ECG reconstruction examples beyond those in the main paper. The displayed windows differ in PPG morphology and motion intensity, while the reconstructed ECG preserves the dominant beat structure and local waveform shape.

\subsection{Additional peak-alignment examples}
\label{app:qual_peaks}

\begin{figure}[!htbp]
    \centering
    \begin{subfigure}[t]{0.32\textwidth}
        \centering
        \includegraphics[width=\textwidth]{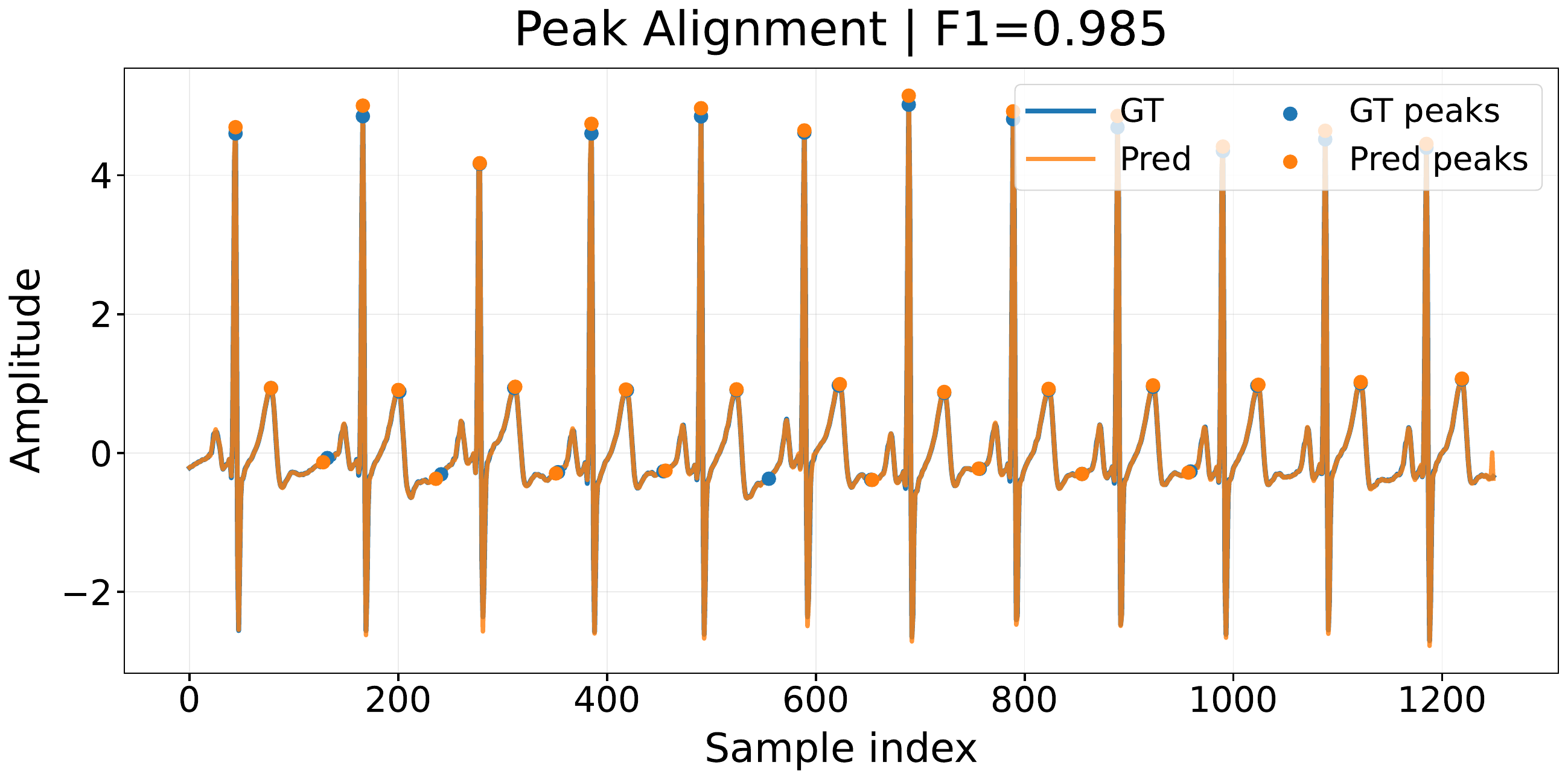}
        \caption{Peak-alignment example 1.}
        \label{fig:app_peak_alignment_main}
    \end{subfigure}
    \hfill
    \begin{subfigure}[t]{0.32\textwidth}
        \centering
        \includegraphics[width=\textwidth]{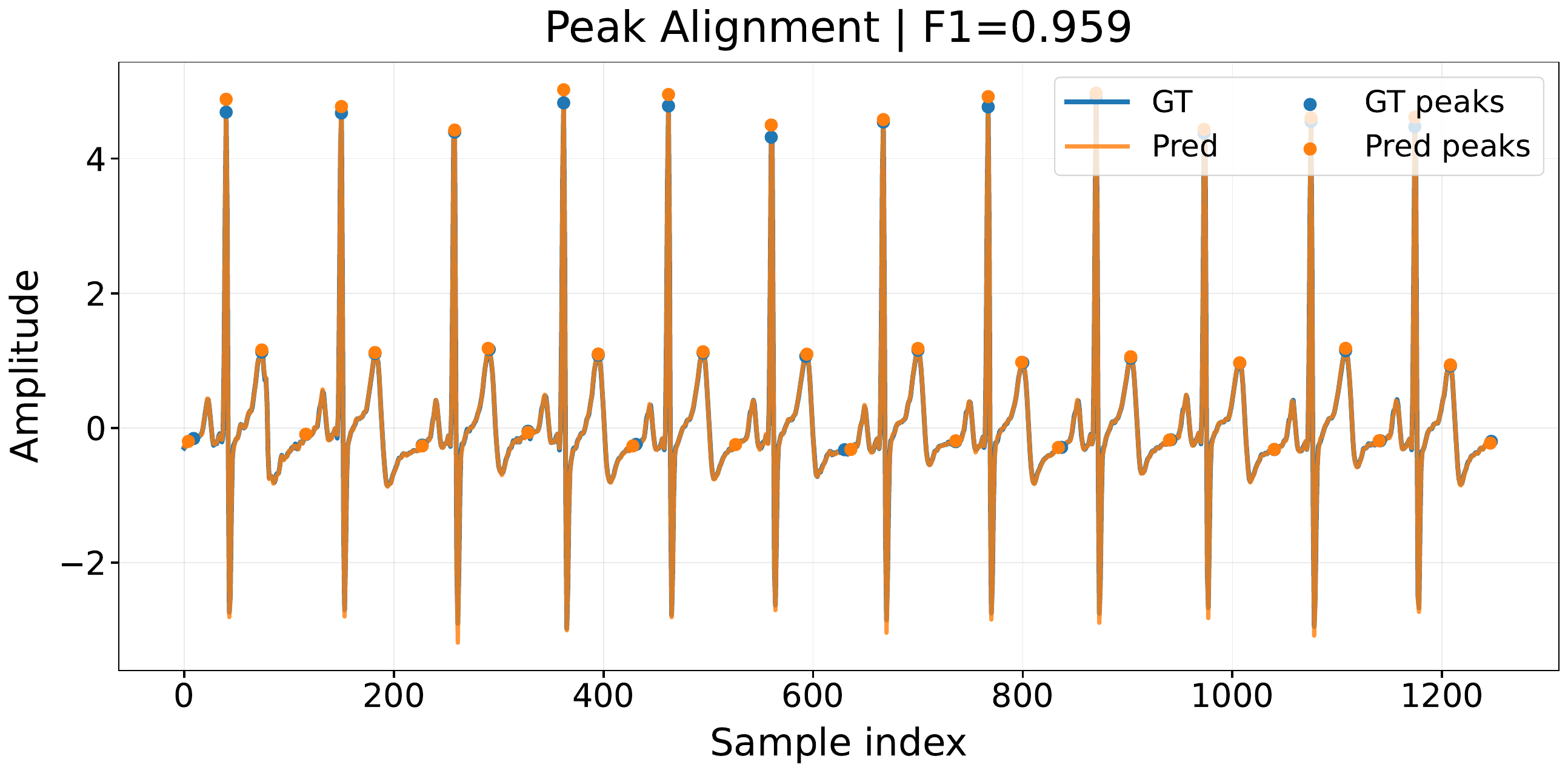}
        \caption{Peak-alignment example 2.}
        \label{fig:app_best_peaks_27299}
    \end{subfigure}
    \hfill
    \begin{subfigure}[t]{0.32\textwidth}
        \centering
        \includegraphics[width=\textwidth]{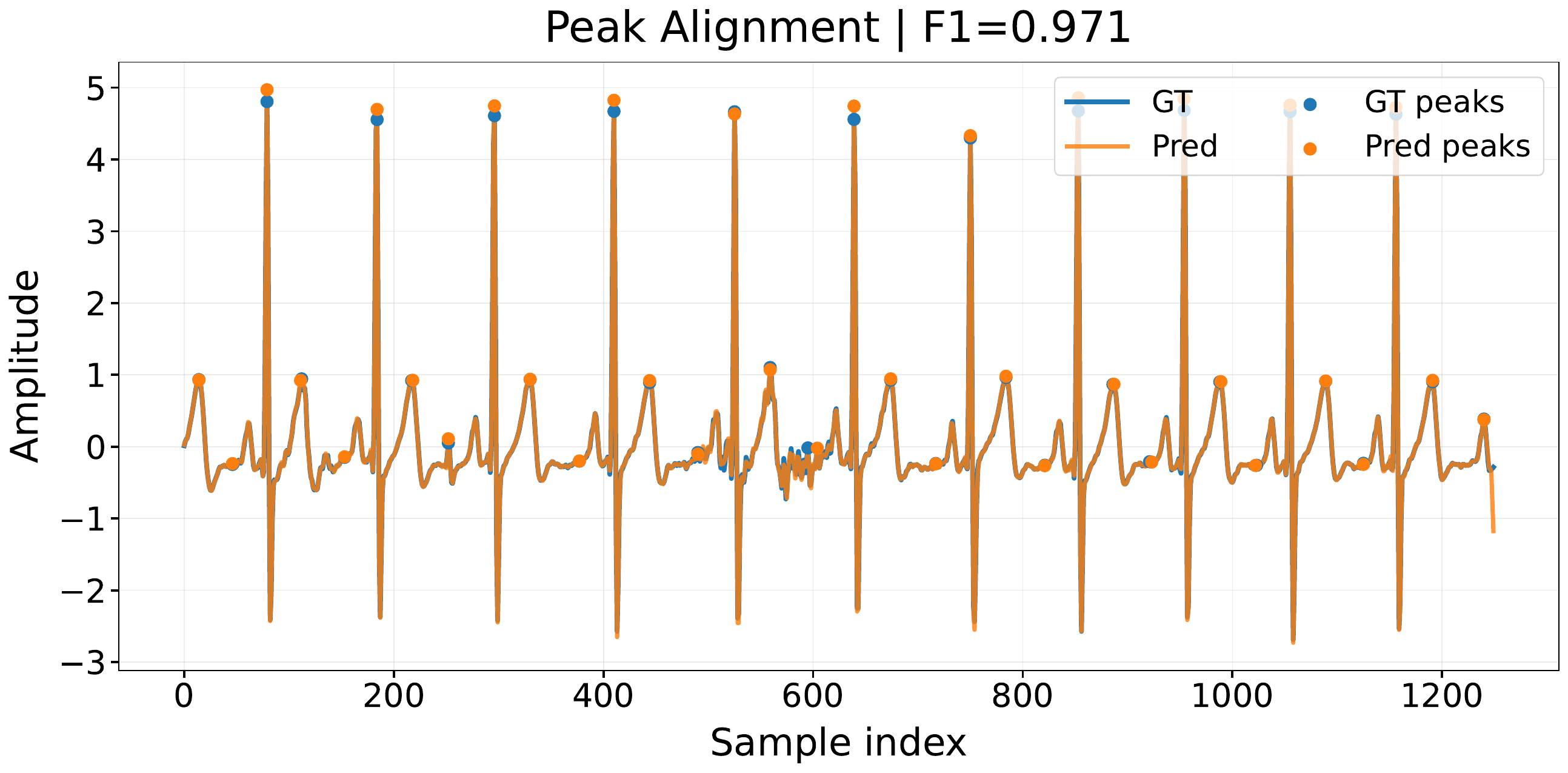}
        \caption{Peak-alignment example 3.}
        \label{fig:app_random_peaks_27545}
    \end{subfigure}

    \caption{\textbf{Peak-alignment examples for BeatFlow-ECG.}
    Predicted and reference R-peaks remain closely aligned across different windows, supporting the R-peak F1 and heart-rate results reported in the main paper.}
    \label{fig:app_peak_examples}
\end{figure}

Figure~\ref{fig:app_peak_examples} shows peak-alignment examples across several windows, including the example referenced in the main paper. Predicted peaks are closely aligned with the target R-peaks in the displayed windows. These examples provide a qualitative view of the beat-timing behavior summarized quantitatively by the R-peak F1 and heart-rate error metrics in the main results.

\subsection{Progressive reconstruction trajectories}
\label{app:qual_traj}

\begin{figure}[!htbp]
    \centering
    \includegraphics[height=0.41\textheight, keepaspectratio]{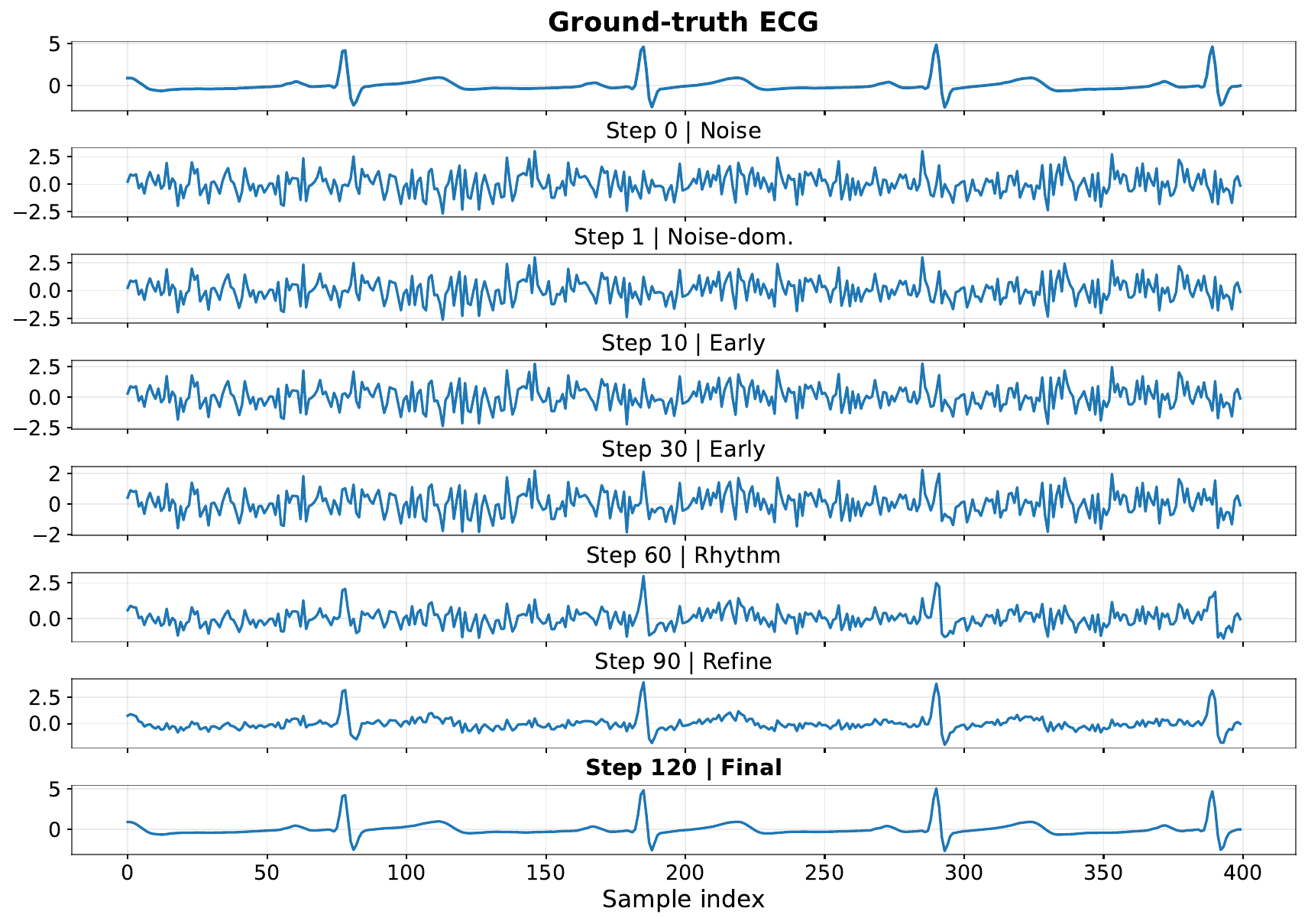}
    \caption{\textbf{Compact progressive reconstruction trajectory.}
    Intermediate states are shown from the initial Gaussian noise sample to the final ECG reconstruction.}
    \label{fig:app_rf_traj_compact}
\end{figure}

\begin{figure}[t]
    \centering
    \includegraphics[width=0.95\textwidth]{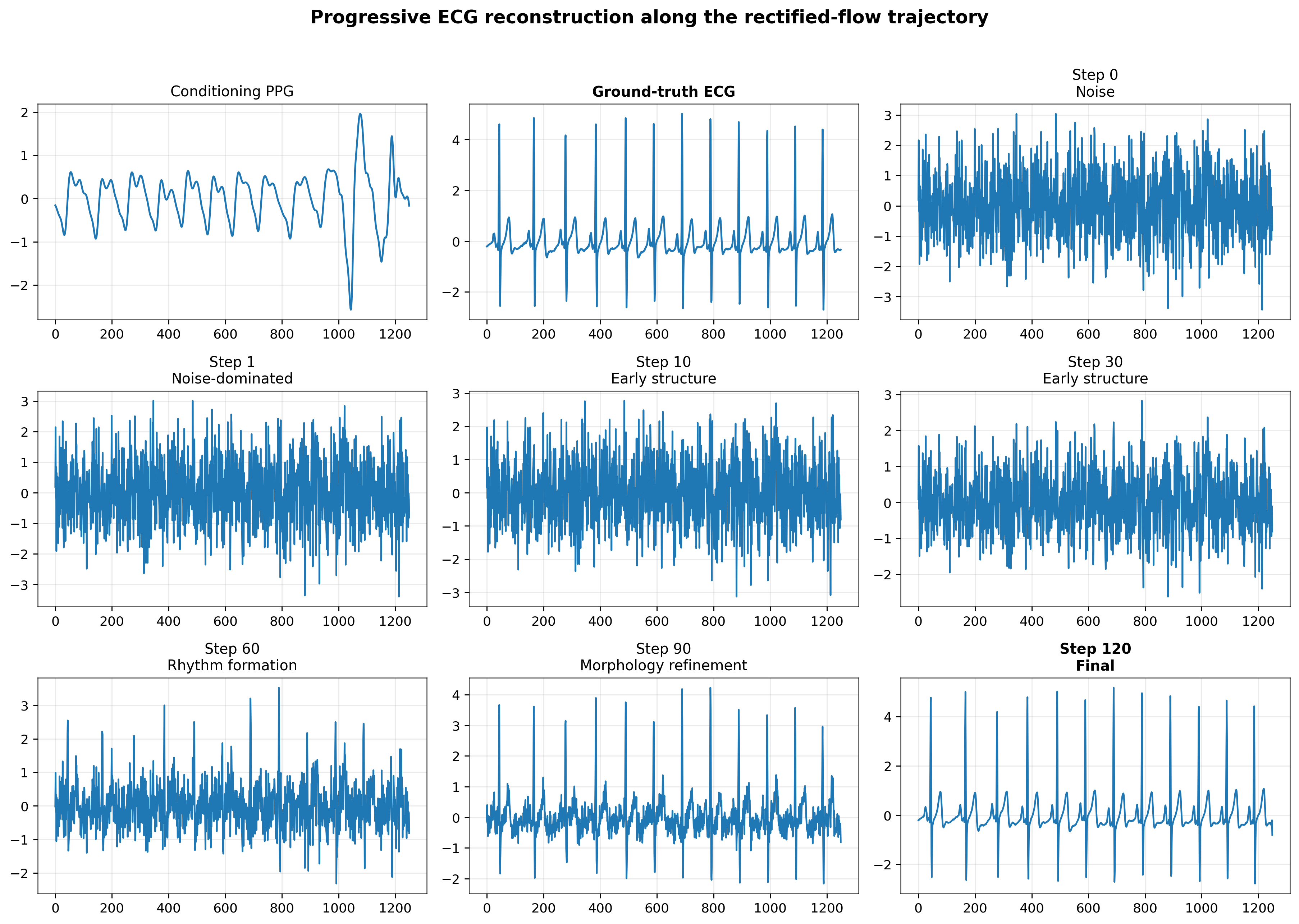}
    \caption{\textbf{Full progressive reconstruction trajectory of BeatFlow-ECG.}
    The figure shows intermediate states along the conditional rectified-flow trajectory from Gaussian noise to the final ECG reconstruction.}
    \label{fig:app_rf_traj_full}
\end{figure}

\begin{figure}[t]
    \centering
    \includegraphics[width=0.95\textwidth]{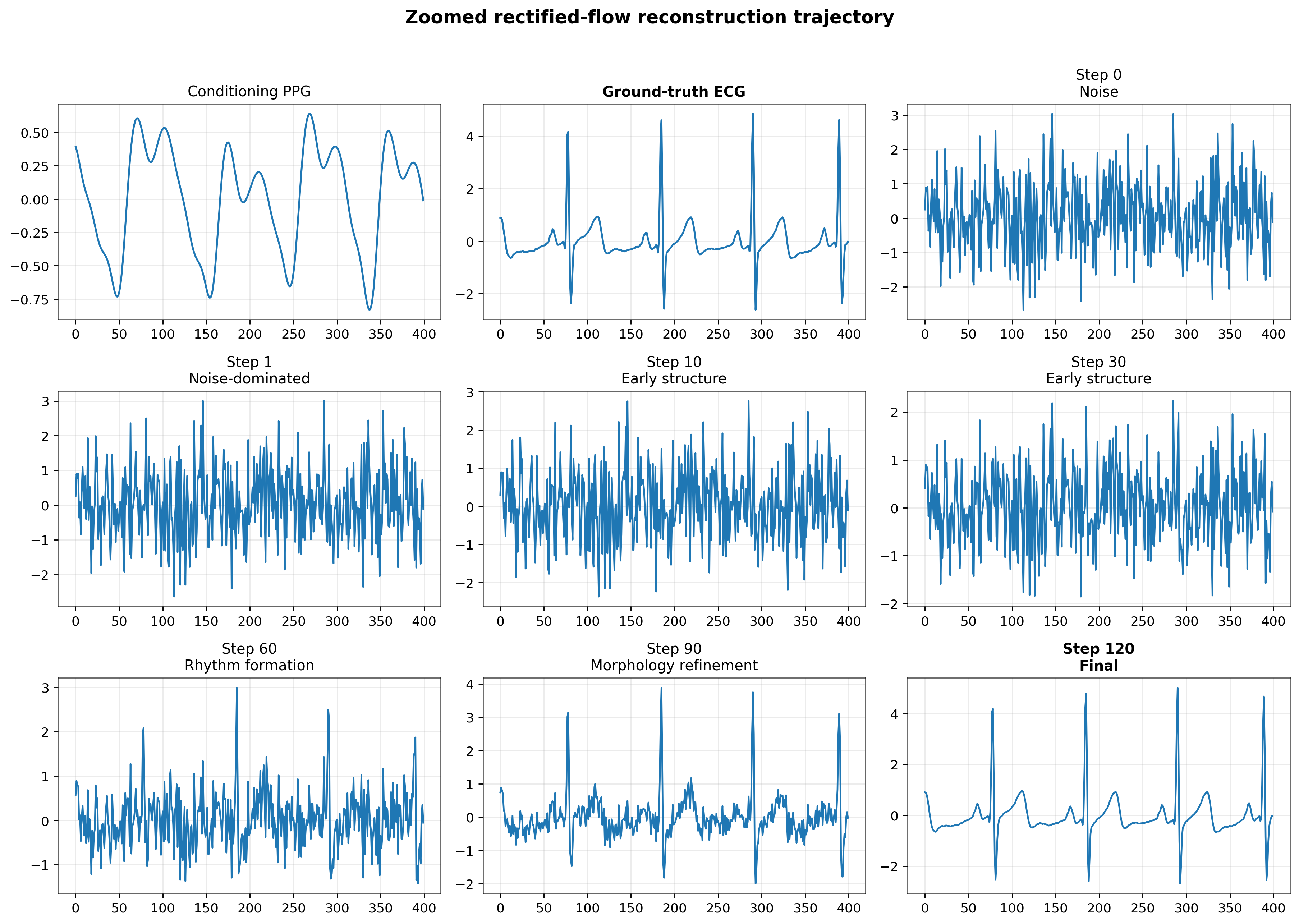}
    \caption{\textbf{Zoomed progressive reconstruction trajectory of BeatFlow-ECG.}
    A shorter temporal segment shows changes in beat timing and waveform structure across intermediate states.}
    \label{fig:app_rf_traj_zoom}
\end{figure}

Figure~\ref{fig:app_rf_traj_compact} shows the compact trajectory view referenced in the main paper, while Figures~\ref{fig:app_rf_traj_full} and~\ref{fig:app_rf_traj_zoom} provide more detailed views of the same reconstruction process. The conditioning signal remains fixed throughout the trajectory, while the intermediate state evolves according to the learned conditional velocity field. The intermediate states show the gradual emergence of ECG structure as the trajectory moves from the initial noise sample toward the final reconstruction. These visualizations are intended to illustrate the iterative reconstruction process rather than define discrete stages of waveform formation.

\FloatBarrier

\FloatBarrier

\section{Additional quantitative results}
\label{app:quant}

\subsection{Inference efficiency}
\label{app:efficiency}

Table~\ref{tab:efficiency_appendix} compares parameter counts, function evaluations, and average inference time for reconstructing a 10-second ECG window (1250 samples) on a single NVIDIA H100 GPU.

\begin{table}[!htbp]
\centering
\caption{\textbf{Inference efficiency of the evaluated models.} Inference time is measured as the average time required to reconstruct a 10-second ECG window (1250 samples) on a single NVIDIA H100 GPU.}
\label{tab:efficiency_appendix}
\resizebox{0.75\textwidth}{!}{%
\begin{tabular}{lccc}
\toprule
\textbf{Model} &
\textbf{Parameters (M)} &
\textbf{Function Evaluations} &
\textbf{Inference Time (ms)} $\downarrow$ \\
\midrule
UNet1D & 4.2 & 1 & 8 \\
AttnSeq & 8.5 & 1 & 14 \\
UNet1D-Large & 13.2 & 1 & 18 \\
AttnSeq-Large & 12.3 & 1 & 22 \\
Conditional DDPM-1D & 12.6 & 1000 & 3520 \\
\textbf{BeatFlow-ECG} & \textbf{12.5} & \textbf{120} & \textbf{425} \\
\bottomrule
\end{tabular}%
}
\end{table}

The deterministic baselines are fastest because they require a single forward pass. BeatFlow-ECG uses iterative integration, but under the evaluated sampling configurations it reduces the number of function evaluations from 1000 to 120 and inference time from 3520 ms to 425 ms relative to Conditional DDPM-1D, corresponding to an approximately 8.3$\times$ speedup. On the evaluated H100 setup, BeatFlow-ECG processes a 10-second input window in 425 ms, substantially shorter than the duration of the signal. This shows that the iterative solver remains computationally practical under the evaluated configuration. The reported timing measures model inference once the input window is available and does not include signal acquisition, preprocessing, or data transfer. Faster diffusion samplers, reduced-step schedules, distillation, or alternative ODE solvers could also change this latency comparison.

\subsection{Motion robustness visualizations}
\label{app:motion_analysis}

\begin{figure}[!htbp]
    \centering
    \begin{subfigure}[t]{0.24\textwidth}
        \centering
        \includegraphics[width=\textwidth]{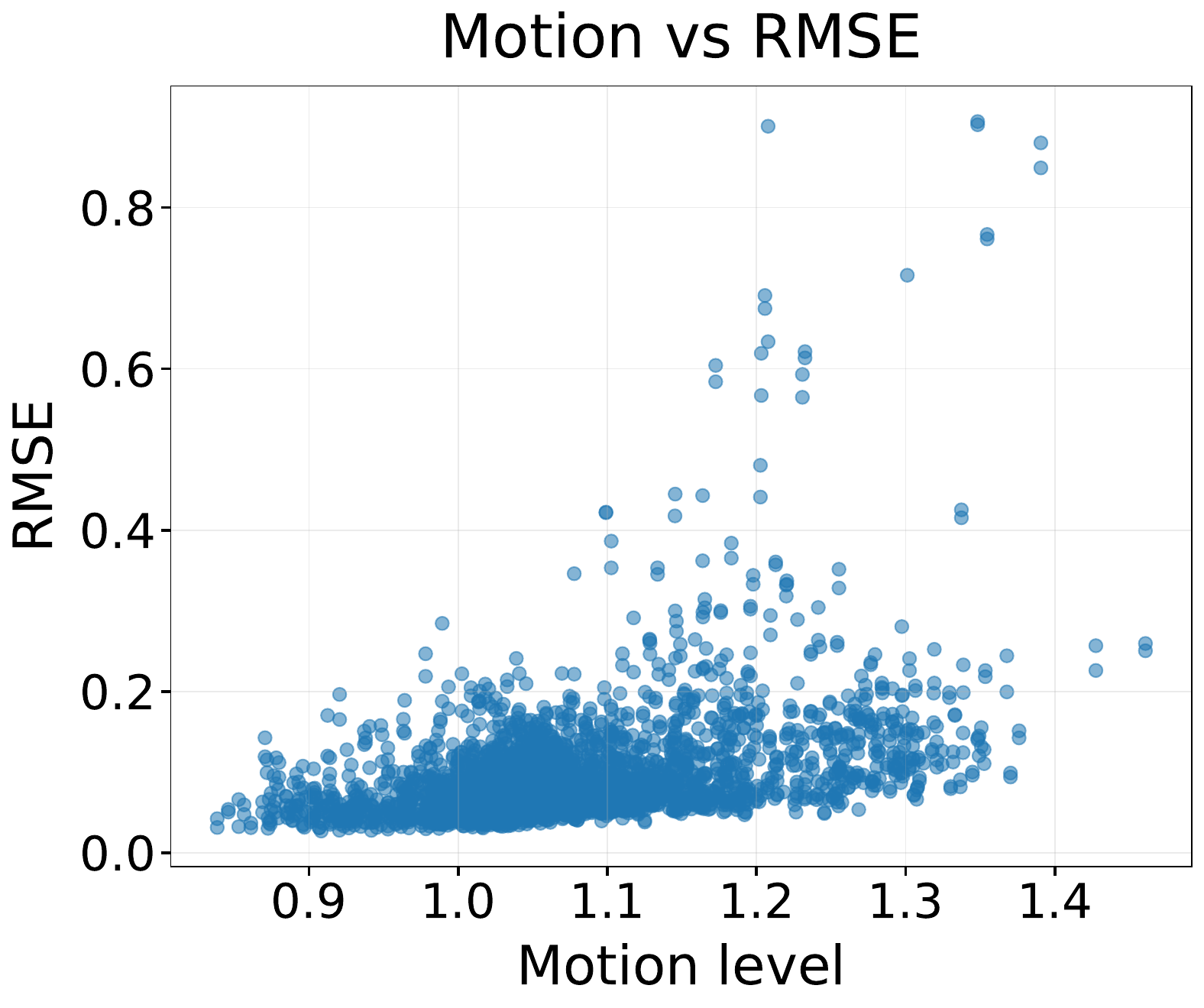}
        \caption{Motion vs RMSE.}
        \label{fig:app_motion_scatter}
    \end{subfigure}\hfill
    \begin{subfigure}[t]{0.24\textwidth}
        \centering
        \includegraphics[width=\textwidth]{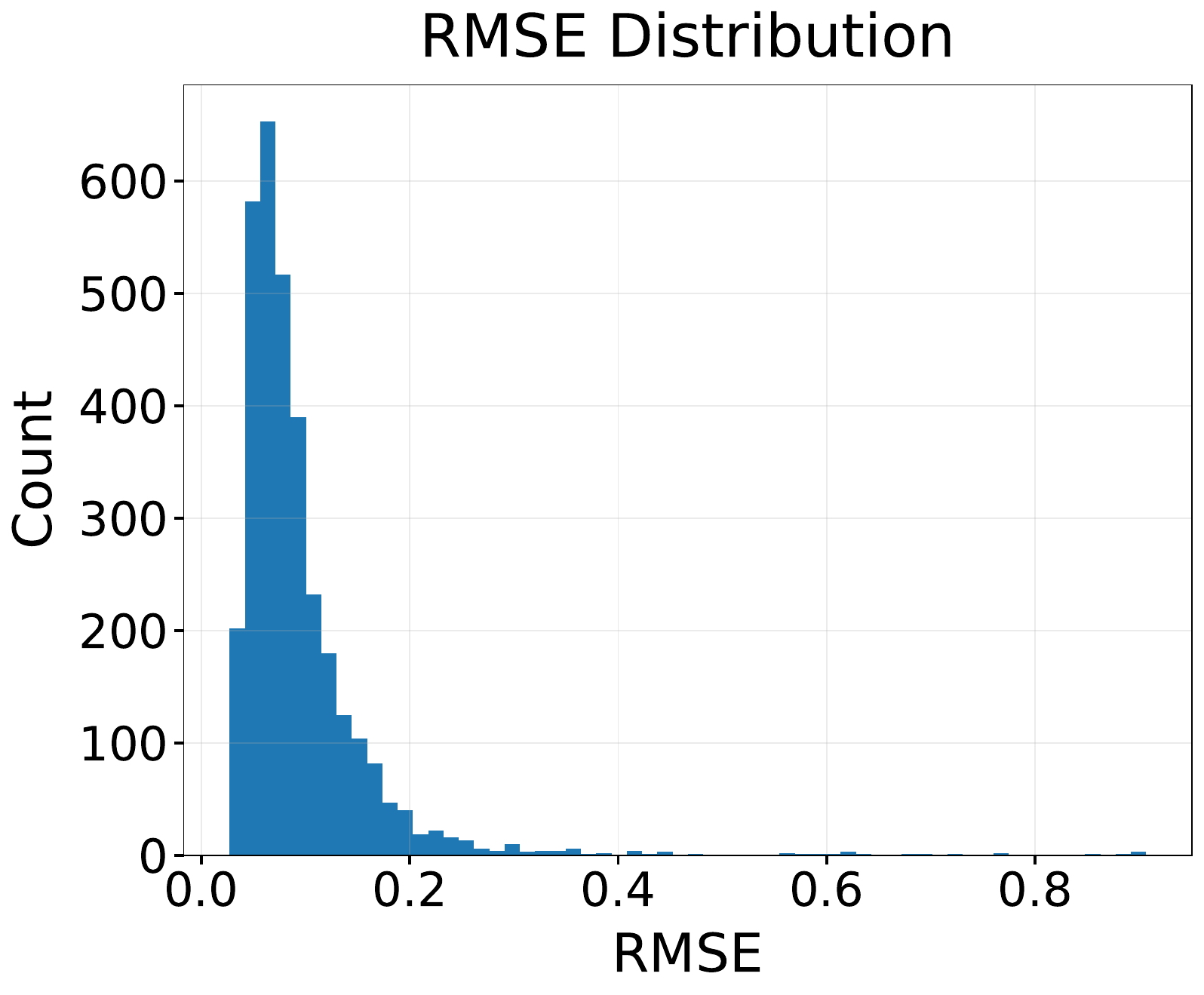}
        \caption{RMSE distribution.}
        \label{fig:app_motion_hist_rmse}
    \end{subfigure}\hfill
    \begin{subfigure}[t]{0.24\textwidth}
        \centering
        \includegraphics[width=\textwidth]{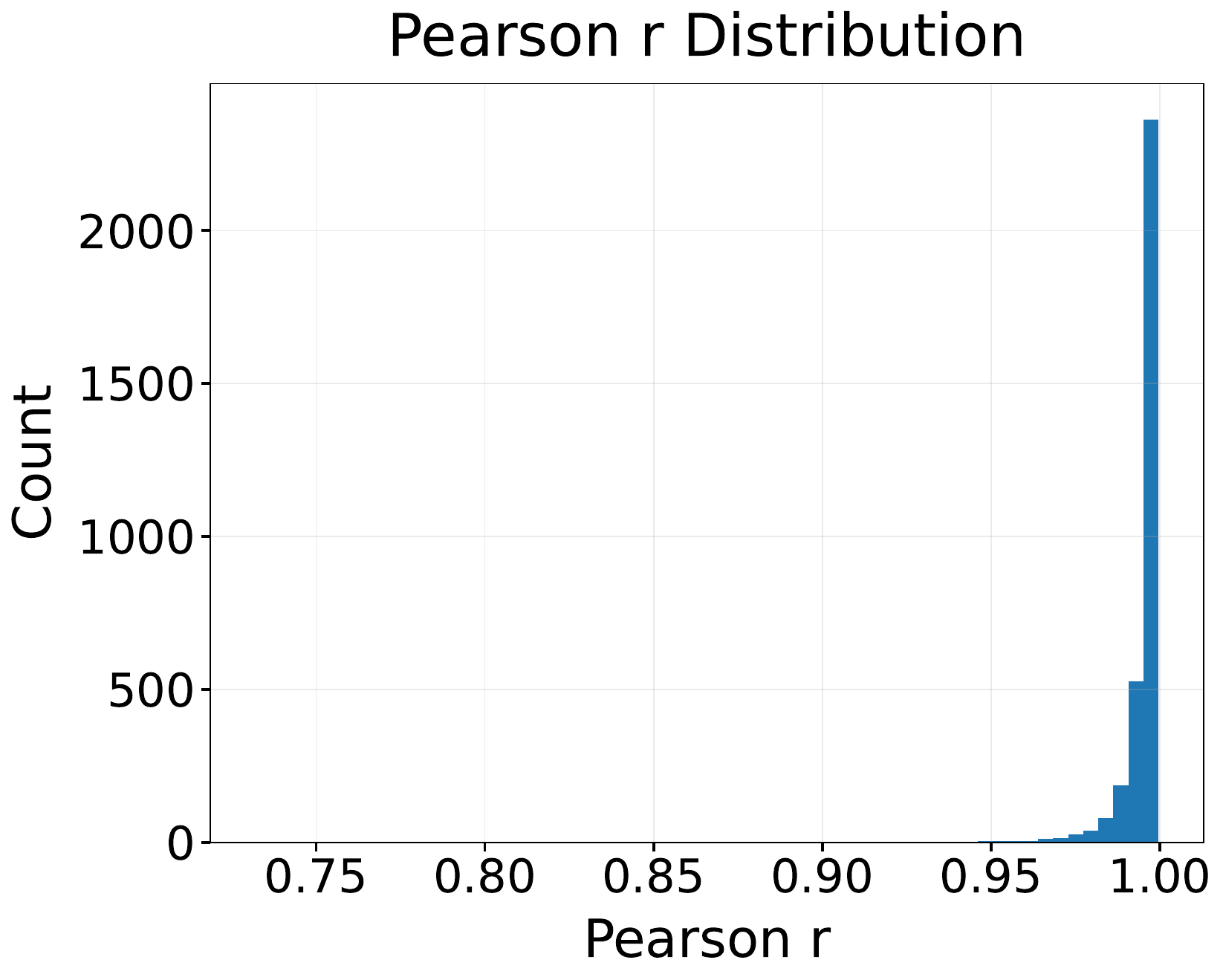}
        \caption{Pearson $r$ dist.}
        \label{fig:app_motion_hist_r}
    \end{subfigure}\hfill
    \begin{subfigure}[t]{0.24\textwidth}
        \centering
        \includegraphics[width=\textwidth]{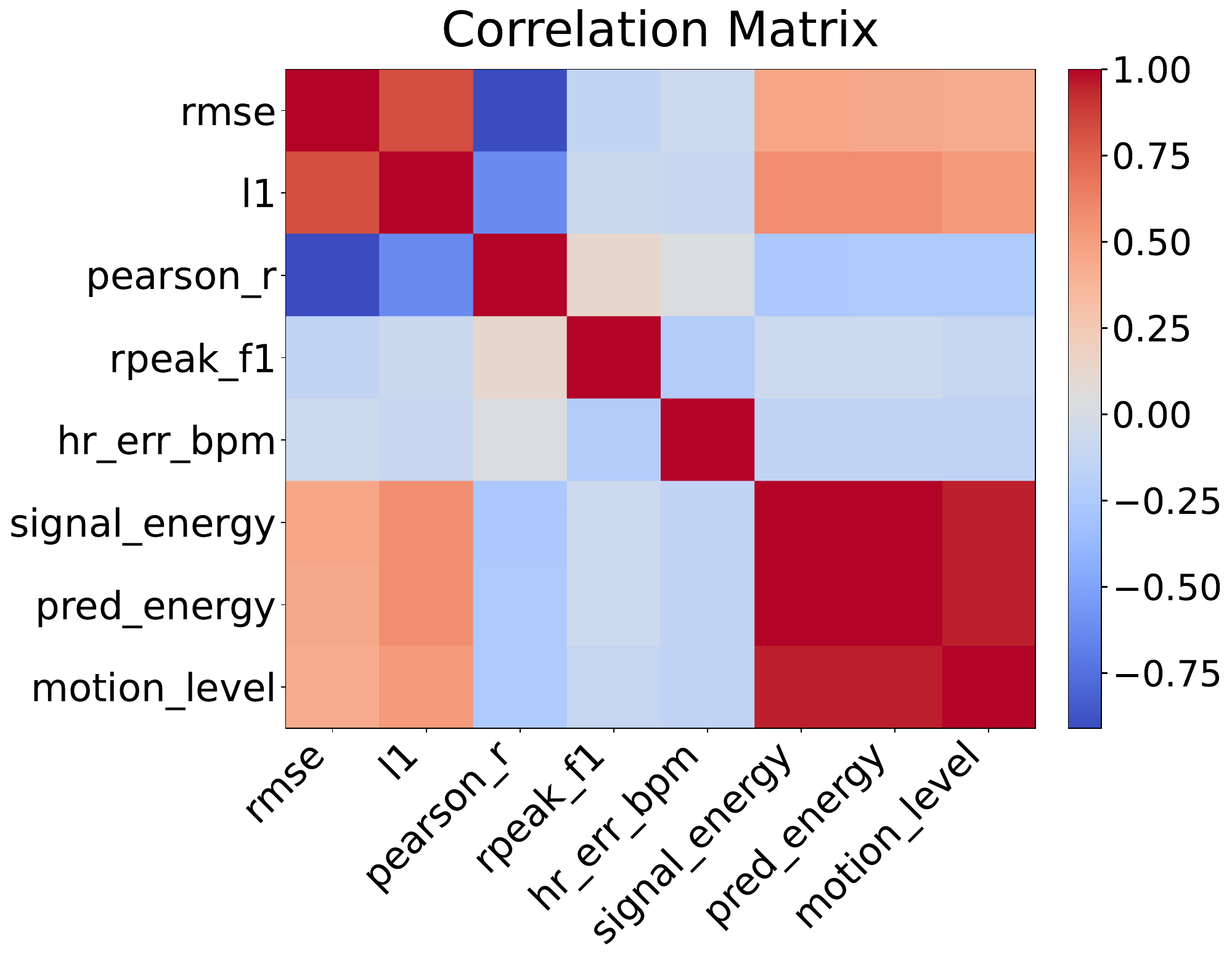}
        \caption{Correlation matrix.}
        \label{fig:app_motion_corr}
    \end{subfigure}

    \caption{\textbf{Motion robustness visualizations.}
    The scatter plot shows reconstruction RMSE against the acceleration-derived motion score. The histograms show the distributions of RMSE and Pearson correlation, and the correlation matrix summarizes relationships among motion level, waveform fidelity, and peak-based metrics.}
    \label{fig:app_motion_analysis}
\end{figure}

Figure~\ref{fig:app_motion_analysis} shows window-level reconstruction metrics and their relationship to the acceleration-derived motion score for BeatFlow-ECG on PPG-DaLiA.

The continuous analysis is consistent with the motion-stratified results in Table~\ref{tab:motion_stratified}: reconstruction error generally increases with the acceleration-derived motion score, although windows at similar motion levels can still exhibit different errors. The distributions further show that most windows are concentrated in the lower-RMSE and higher-correlation regions, providing additional context beyond the three motion groups used in the main paper.

\FloatBarrier

\subsection{Per-subject tables}
\label{app:quant_subjects}

The following tables report subject-wise LOSO results for BeatFlow-ECG, Conditional DDPM-1D, and UNet1D-Large on PPG-DaLiA and WESAD. Reporting individual held-out folds makes it possible to examine cross-subject variability that is not visible from aggregate means alone and to determine whether the overall results are driven by a small subset of subjects.

On PPG-DaLiA, S6 and S10 remain among the more difficult folds across the evaluated models, consistent with the subject-wise discussion in the main paper. BeatFlow-ECG nevertheless achieves lower reconstruction error than Conditional DDPM-1D and UNet1D-Large on these difficult subjects. The complete tables also show variation in waveform, peak-detection, and heart-rate metrics across all held-out subjects on both datasets.


\begin{table}[!htbp]
    \centering
    \renewcommand{\arraystretch}{0.88}
    \caption{\textbf{Subject-wise LOSO performance of BeatFlow-ECG across 15 subjects on PPG-DaLiA.}}
    \label{tab:app_subjectwise_beatflow_dalia}
    \resizebox{0.85\textwidth}{!}{%
    \begin{tabular}{lccccc}
    \toprule
    \textbf{Subject} & \textbf{L1} $\downarrow$ & \textbf{RMSE} $\downarrow$ & \textbf{Pearson $r$} $\uparrow$ & \textbf{R-peak F1} $\uparrow$ & \textbf{HR error (bpm)} $\downarrow$ \\
    \midrule
    S1  & 0.063 & 0.106 & 0.986 & 0.943 & 5.14 \\
    S2  & 0.041 & 0.075 & 0.996 & 0.965 & 4.15 \\
    S3  & 0.065 & 0.102 & 0.989 & 0.947 & 6.46 \\
    S4  & 0.044 & 0.080 & 0.995 & 0.958 & 4.37 \\
    S5  & 0.045 & 0.081 & 0.995 & 0.975 & 3.85 \\
    S6  & 0.097 & 0.143 & 0.915 & 0.885 & 6.55 \\
    S7  & 0.047 & 0.087 & 0.993 & 0.955 & 4.21 \\
    S8  & 0.071 & 0.127 & 0.976 & 0.945 & 4.88 \\
    S9  & 0.045 & 0.081 & 0.995 & 0.964 & 4.10 \\
    S10 & 0.098 & 0.145 & 0.964 & 0.926 & 4.45 \\
    S11 & 0.045 & 0.083 & 0.994 & 0.964 & 4.37 \\
    S12 & 0.079 & 0.122 & 0.982 & 0.927 & 5.67 \\
    S13 & 0.044 & 0.084 & 0.995 & 0.962 & 4.38 \\
    S14 & 0.078 & 0.119 & 0.979 & 0.928 & 5.00 \\
    S15 & 0.063 & 0.101 & 0.989 & 0.938 & 5.78 \\
    \midrule
    \textbf{Overall} & \textbf{0.062} & \textbf{0.102} & \textbf{0.983} & \textbf{0.946} & \textbf{4.89} \\
    \bottomrule
    \end{tabular}%
    }
\end{table}

\begin{table}[!htbp]
    \centering
    \renewcommand{\arraystretch}{0.88}
    \caption{\textbf{Subject-wise LOSO performance of Conditional DDPM-1D across 15 subjects on PPG-DaLiA.}}
    \label{tab:app_subjectwise_ddpm_dalia}
    \resizebox{0.85\textwidth}{!}{%
    \begin{tabular}{lccccc}
    \toprule
    \textbf{Subject} & \textbf{L1} $\downarrow$ & \textbf{RMSE} $\downarrow$ & \textbf{Pearson $r$} $\uparrow$ & \textbf{R-peak F1} $\uparrow$ & \textbf{HR error (bpm)} $\downarrow$ \\
    \midrule
    S1  & 0.086 & 0.132 & 0.967 & 0.912 & 6.20 \\
    S2  & 0.052 & 0.092 & 0.988 & 0.942 & 4.90 \\
    S3  & 0.088 & 0.130 & 0.972 & 0.920 & 7.30 \\
    S4  & 0.055 & 0.098 & 0.985 & 0.935 & 5.10 \\
    S5  & 0.058 & 0.101 & 0.985 & 0.948 & 4.70 \\
    S6  & 0.142 & 0.202 & 0.880 & 0.805 & 8.20 \\
    S7  & 0.060 & 0.105 & 0.982 & 0.928 & 5.00 \\
    S8  & 0.108 & 0.168 & 0.945 & 0.885 & 6.90 \\
    S9  & 0.056 & 0.099 & 0.985 & 0.938 & 4.80 \\
    S10 & 0.145 & 0.208 & 0.920 & 0.852 & 6.30 \\
    S11 & 0.058 & 0.102 & 0.983 & 0.938 & 5.00 \\
    S12 & 0.112 & 0.162 & 0.952 & 0.875 & 7.20 \\
    S13 & 0.055 & 0.101 & 0.986 & 0.938 & 5.10 \\
    S14 & 0.110 & 0.158 & 0.948 & 0.872 & 6.80 \\
    S15 & 0.090 & 0.122 & 0.962 & 0.887 & 8.30 \\
    \midrule
    \textbf{Overall} & \textbf{0.085} & \textbf{0.132} & \textbf{0.962} & \textbf{0.905} & \textbf{6.12} \\
    \bottomrule
    \end{tabular}%
    }
\end{table}

\begin{table}[!htbp]
    \centering
    \renewcommand{\arraystretch}{0.88}
    \caption{\textbf{Subject-wise LOSO performance of UNet1D-Large across 15 subjects on PPG-DaLiA.}}
    \label{tab:app_subjectwise_unet_large_dalia}
    \resizebox{0.85\textwidth}{!}{%
    \begin{tabular}{lccccc}
    \toprule
    \textbf{Subject} & \textbf{L1} $\downarrow$ & \textbf{RMSE} $\downarrow$ & \textbf{Pearson $r$} $\uparrow$ & \textbf{R-peak F1} $\uparrow$ & \textbf{HR error (bpm)} $\downarrow$ \\
    \midrule
    S1  & 0.175 & 0.279 & 0.925 & 0.840 & 7.05 \\
    S2  & 0.035 & 0.057 & 0.998 & 0.950 & 5.75 \\
    S3  & 0.155 & 0.220 & 0.948 & 0.865 & 8.20 \\
    S4  & 0.048 & 0.079 & 0.995 & 0.930 & 5.50 \\
    S5  & 0.056 & 0.096 & 0.993 & 0.980 & 2.98 \\
    S6  & 0.282 & 0.396 & 0.655 & 0.685 & 10.65 \\
    S7  & 0.056 & 0.102 & 0.988 & 0.920 & 5.56 \\
    S8  & 0.218 & 0.362 & 0.885 & 0.835 & 9.90 \\
    S9  & 0.054 & 0.094 & 0.993 & 0.945 & 4.45 \\
    S10 & 0.281 & 0.405 & 0.825 & 0.805 & 5.58 \\
    S11 & 0.050 & 0.094 & 0.989 & 0.942 & 5.98 \\
    S12 & 0.218 & 0.315 & 0.915 & 0.810 & 10.15 \\
    S13 & 0.043 & 0.086 & 0.995 & 0.935 & 5.18 \\
    S14 & 0.239 & 0.340 & 0.820 & 0.835 & 8.10 \\
    S15 & 0.160 & 0.225 & 0.951 & 0.848 & 9.22 \\
    \midrule
    \textbf{Overall} & \textbf{0.138} & \textbf{0.210} & \textbf{0.925} & \textbf{0.875} & \textbf{6.95} \\
    \bottomrule
    \end{tabular}%
    }
\end{table}


\begin{table}[!htbp]
    \centering
    \renewcommand{\arraystretch}{0.88}
    \caption{\textbf{Subject-wise LOSO performance of BeatFlow-ECG across 15 subjects on WESAD.}}
    \label{tab:app_subjectwise_beatflow_wesad}
    \resizebox{0.85\textwidth}{!}{%
    \begin{tabular}{lccccc}
    \toprule
    \textbf{Subject} & \textbf{L1} $\downarrow$ & \textbf{RMSE} $\downarrow$ & \textbf{Pearson $r$} $\uparrow$ & \textbf{R-peak F1} $\uparrow$ & \textbf{HR error (bpm)} $\downarrow$ \\
    \midrule
    S1  & 0.056 & 0.091 & 0.988 & 0.952 & 4.25 \\
    S2  & 0.038 & 0.068 & 0.997 & 0.971 & 3.50 \\
    S3  & 0.058 & 0.094 & 0.990 & 0.958 & 5.10 \\
    S4  & 0.040 & 0.072 & 0.996 & 0.965 & 3.75 \\
    S5  & 0.042 & 0.075 & 0.996 & 0.980 & 3.20 \\
    S6  & 0.082 & 0.125 & 0.935 & 0.905 & 5.50 \\
    S7  & 0.041 & 0.078 & 0.995 & 0.962 & 3.65 \\
    S8  & 0.064 & 0.110 & 0.982 & 0.950 & 4.15 \\
    S9  & 0.039 & 0.070 & 0.996 & 0.970 & 3.45 \\
    S10 & 0.088 & 0.130 & 0.975 & 0.940 & 3.90 \\
    S11 & 0.042 & 0.074 & 0.995 & 0.972 & 3.60 \\
    S12 & 0.070 & 0.105 & 0.985 & 0.938 & 4.80 \\
    S13 & 0.040 & 0.076 & 0.996 & 0.968 & 3.70 \\
    S14 & 0.068 & 0.108 & 0.981 & 0.942 & 4.20 \\
    S15 & 0.057 & 0.089 & 0.990 & 0.948 & 5.05 \\
    \midrule
\textbf{Overall} & \textbf{0.055} & \textbf{0.091} & \textbf{0.986} & \textbf{0.955} & \textbf{4.12} \\
    \bottomrule
    \end{tabular}%
    }
\end{table}

\begin{table}[!htbp]
    \centering
    \renewcommand{\arraystretch}{0.88}
    \caption{\textbf{Subject-wise LOSO performance of Conditional DDPM-1D across 15 subjects on WESAD.}}
    \label{tab:app_subjectwise_ddpm_wesad}
    \resizebox{0.85\textwidth}{!}{%
    \begin{tabular}{lccccc}
    \toprule
    \textbf{Subject} & \textbf{L1} $\downarrow$ & \textbf{RMSE} $\downarrow$ & \textbf{Pearson $r$} $\uparrow$ & \textbf{R-peak F1} $\uparrow$ & \textbf{HR error (bpm)} $\downarrow$ \\
    \midrule
    S1  & 0.076 & 0.114 & 0.974 & 0.915 & 5.40 \\
    S2  & 0.048 & 0.082 & 0.991 & 0.952 & 4.20 \\
    S3  & 0.078 & 0.118 & 0.977 & 0.922 & 6.10 \\
    S4  & 0.052 & 0.088 & 0.988 & 0.940 & 4.50 \\
    S5  & 0.054 & 0.091 & 0.989 & 0.955 & 3.90 \\
    S6  & 0.112 & 0.160 & 0.907 & 0.835 & 6.80 \\
    S7  & 0.053 & 0.092 & 0.986 & 0.938 & 4.30 \\
    S8  & 0.088 & 0.134 & 0.958 & 0.902 & 5.20 \\
    S9  & 0.050 & 0.085 & 0.990 & 0.948 & 4.10 \\
    S10 & 0.118 & 0.166 & 0.944 & 0.875 & 4.90 \\
    S11 & 0.052 & 0.089 & 0.988 & 0.948 & 4.40 \\
    S12 & 0.092 & 0.132 & 0.962 & 0.885 & 5.90 \\
    S13 & 0.051 & 0.089 & 0.989 & 0.942 & 4.40 \\
    S14 & 0.090 & 0.130 & 0.955 & 0.888 & 5.20 \\
    S15 & 0.097 & 0.130 & 0.967 & 0.925 & 9.45 \\
    \midrule
\textbf{Overall} & \textbf{0.074} & \textbf{0.113} & \textbf{0.971} & \textbf{0.918} & \textbf{5.25} \\
    \bottomrule
    \end{tabular}%
    }
\end{table}

\begin{table}[!htbp]
    \centering
    \renewcommand{\arraystretch}{0.88}
    \caption{\textbf{Subject-wise LOSO performance of UNet1D-Large across 15 subjects on WESAD.}}
    \label{tab:app_subjectwise_unet_large_wesad}
    \resizebox{0.85\textwidth}{!}{%
    \begin{tabular}{lccccc}
    \toprule
    \textbf{Subject} & \textbf{L1} $\downarrow$ & \textbf{RMSE} $\downarrow$ & \textbf{Pearson $r$} $\uparrow$ & \textbf{R-peak F1} $\uparrow$ & \textbf{HR error (bpm)} $\downarrow$ \\
    \midrule
    S1  & 0.163 & 0.258 & 0.918 & 0.862 & 6.55 \\
    S2  & 0.038 & 0.058 & 0.998 & 0.955 & 5.35 \\
    S3  & 0.142 & 0.205 & 0.952 & 0.888 & 7.60 \\
    S4  & 0.048 & 0.078 & 0.995 & 0.932 & 5.10 \\
    S5  & 0.055 & 0.092 & 0.994 & 0.981 & 2.80 \\
    S6  & 0.260 & 0.355 & 0.550 & 0.655 & 9.85 \\
    S7  & 0.055 & 0.098 & 0.989 & 0.922 & 5.15 \\
    S8  & 0.202 & 0.325 & 0.882 & 0.850 & 9.18 \\
    S9  & 0.052 & 0.091 & 0.994 & 0.952 & 4.10 \\
    S10 & 0.258 & 0.365 & 0.812 & 0.805 & 5.18 \\
    S11 & 0.048 & 0.090 & 0.990 & 0.948 & 5.55 \\
    S12 & 0.201 & 0.290 & 0.915 & 0.802 & 9.42 \\
    S13 & 0.042 & 0.082 & 0.997 & 0.945 & 4.80 \\
    S14 & 0.222 & 0.313 & 0.882 & 0.802 & 7.50 \\
    S15 & 0.134 & 0.210 & 0.950 & 0.841 & 8.17 \\
    \midrule
    \textbf{Overall} & \textbf{0.128} & \textbf{0.194} & \textbf{0.921} & \textbf{0.876} & \textbf{6.42} \\
    \bottomrule
    \end{tabular}%
    }
\end{table}
\FloatBarrier



\subsection{Baseline qualitative comparisons}
\label{app:quant_baselines}

\begin{figure}[!htbp]
    \centering
    \begin{subfigure}[t]{0.48\textwidth}
        \centering
        \includegraphics[width=\textwidth]{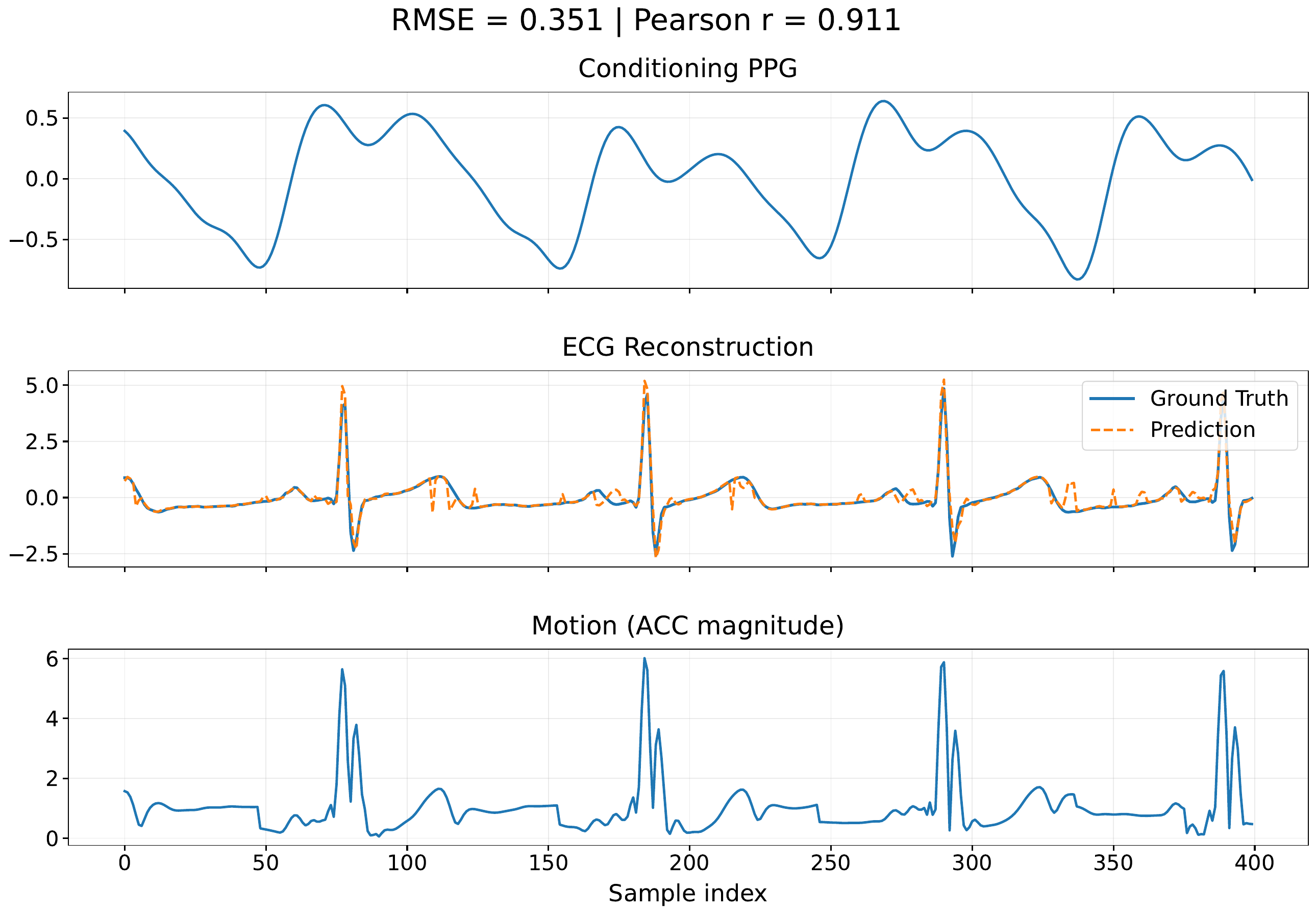}
        \caption{AttnSeq reconstruction on a selected test window.}
        \label{fig:app_attn_zoom}
    \end{subfigure}
    \hfill
    \begin{subfigure}[t]{0.48\textwidth}
        \centering
        \includegraphics[width=\textwidth]{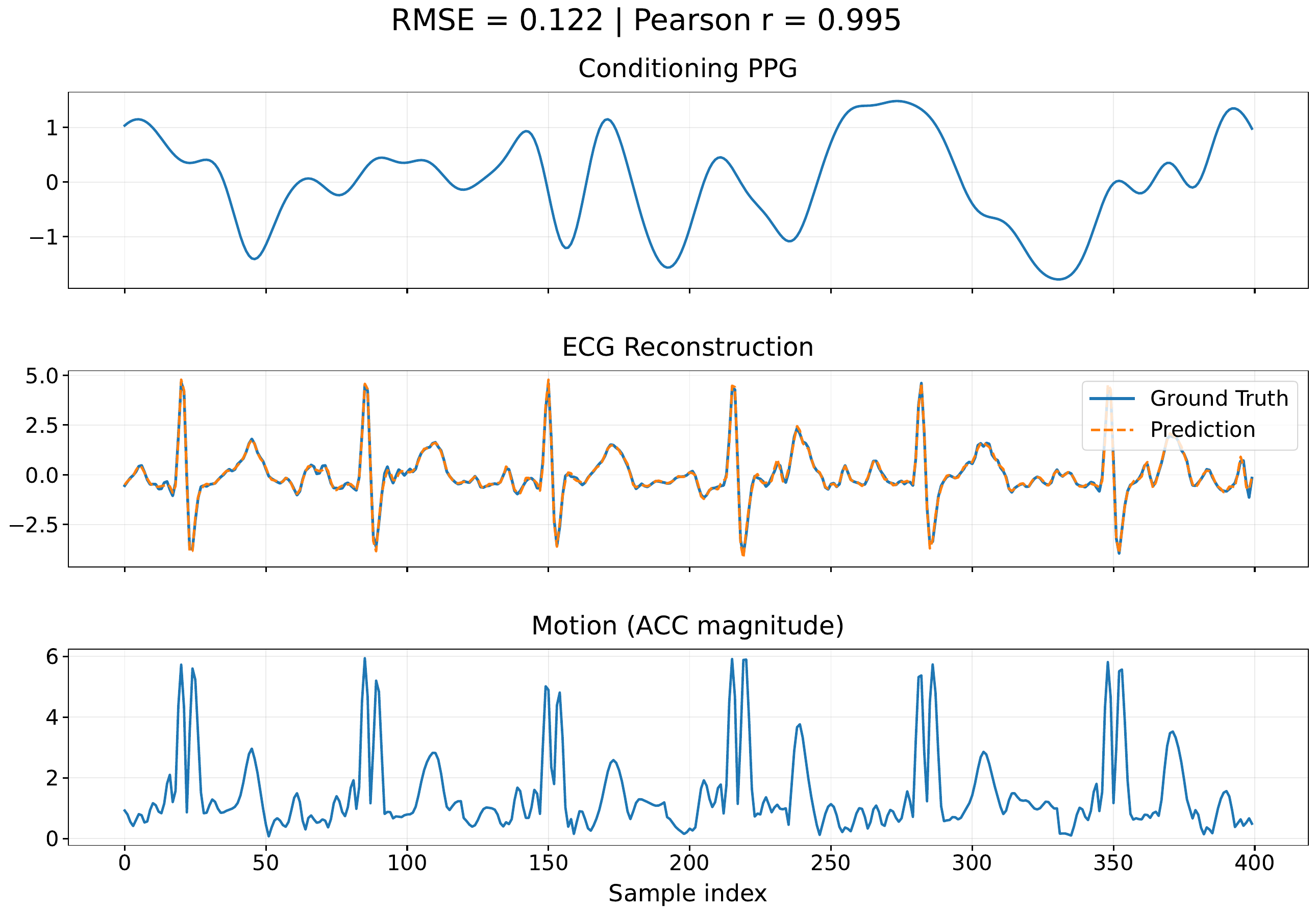}
        \caption{UNet1D reconstruction on a selected test window.}
        \label{fig:app_unet_zoom}
    \end{subfigure}
    \caption{\textbf{Additional baseline qualitative comparisons.}
    AttnSeq and UNet1D reconstructions on selected test windows.}
    \label{fig:app_baseline_compare}
\end{figure}

Figure~\ref{fig:app_baseline_compare} provides additional qualitative examples from the supervised baselines. The displayed windows show that these models can reconstruct some ECG segments closely, while other examples contain local morphology, amplitude, or temporal-alignment errors. These examples complement the aggregate results in Table~\ref{tab:overall_comparison} and the subject-wise results in Appendix~\ref{app:quant_subjects}.

\subsection{Morphology-sensitive region analysis}
\label{app:segment_analysis}

For the morphology-sensitive evaluation reported in Table~\ref{tab:morphology_main}, we use fixed R-peak-centered timing regions as proxies for the P-wave, QRS complex, and T-wave. Around each reference R-peak, we consider a 600-ms interval and define the P-wave region as $-200$ to $-60$ ms, the QRS region as $-60$ to $+60$ ms, and the T-wave region as $+60$ to $+400$ ms relative to the R-peak. Pearson correlation between the reconstructed and reference ECG is computed independently within each region. Reference R-peaks are used to define the same temporal regions for all models.

\section{Reproducibility details}
\label{app:repro}

\subsection{Baseline model details}
\label{app:baseline_details}

We benchmark BeatFlow-ECG against deterministic, adversarial, and diffusion-based baselines. UNet1D, AttnSeq, and Conditional DDPM-1D use the same conditioning tensor $c \in \mathbb{R}^{C \times T}$, target ECG waveform $x \in \mathbb{R}^{1 \times T}$, LOSO splits, preprocessing pipeline, and evaluation metrics as BeatFlow-ECG. CardioGAN uses PPG only, following its original formulation, and is evaluated under the same dataset-specific LOSO protocol, preprocessing pipeline, and evaluation metrics.

\paragraph{UNet-based regressors (UNet1D and UNet1D-Large).}
UNet1D is a 1D convolutional encoder--decoder architecture. The standard variant (\textbf{UNet1D}, 4.2M parameters) projects the multimodal conditioning input using a Conv-BatchNorm-GELU stem (base width 64, kernel size 9), followed by three downsampling stages with channel widths 128, 256, and 512 using max pooling. For a capacity-matched comparison, we also construct a larger variant (\textbf{UNet1D-Large}, 13.2M parameters) by expanding the base channel width to 128 and increasing intermediate block depths (channel widths 256, 512, and 1024), yielding a parameter count comparable to BeatFlow-ECG.

\paragraph{Attention-based regressors (AttnSeq and AttnSeq-Large).}
AttnSeq is an encoder-only transformer baseline for direct waveform regression. The standard variant (\textbf{AttnSeq}, 8.5M parameters) projects the input to a latent dimension of 128 and processes it through a 4-layer Transformer encoder with 4 attention heads and feedforward dimension 256. The capacity-matched variant (\textbf{AttnSeq-Large}, 12.3M parameters) expands the latent projection to 256, increases the depth to 8 layers with 8 attention heads, and sets the feedforward dimension to 1024.

\paragraph{Conditional DDPM-1D.}
We implement a 1D conditional Denoising Diffusion Probabilistic Model (\textbf{Conditional DDPM-1D}, 12.6M parameters) inspired by BeatDiff \citep{NEURIPS2024_9988f2c8}. The model uses the same hybrid convolutional-transformer backbone and conditioning inputs as BeatFlow-ECG, but is trained with a 1000-step linear noise schedule $\beta_t \in [10^{-4}, 0.02]$. At inference, reconstruction is performed by reverse diffusion over 1000 timesteps starting from standard Gaussian noise.

\paragraph{CardioGAN.}
We also include CardioGAN \citep{Sarkar_Etemad_2021} as a published PPG-to-ECG generative baseline. The original CardioGAN study combines BIDMC, CAPNO, PPG-DaLiA, and WESAD and uses a different user split and preprocessing setup from our LOSO evaluation. The released pretrained weights were trained on pooled data that include PPG-DaLiA and WESAD, so we do not use them directly because they may contain information from subjects held out in our LOSO folds.

For CardioGAN, we use PPG only and retain the original two-generator, four-discriminator architecture with time- and frequency-domain discriminators, adversarial loss, and cycle-consistency loss. We train the model under the same dataset-specific LOSO protocol and preprocessing pipeline used in our evaluation, using 10-second windows resampled to 125 Hz. Training uses Adam with learning rate $10^{-4}$, batch size 128, and 15 epochs, with learning-rate decay after epoch 10 and loss weights $\alpha=3$, $\beta=1$, and $\lambda=30$. The generators used at inference contain 28.24M parameters, while the complete training system including the discriminators contains 78.92M parameters. The CardioGAN results reported in Table~\ref{tab:overall_comparison} are obtained under our subject-independent LOSO evaluation on PPG-DaLiA and WESAD. Accordingly, these results should be interpreted as CardioGAN evaluated under our LOSO protocol rather than as a reproduction of the original paper's reported experimental setting.

\paragraph{Training objective and protocol.}
The deterministic baselines are trained using an $L_1$ reconstruction loss with AdamW (learning rate $10^{-3}$, batch size 64, 30 epochs, and cosine annealing). The final PPG-DaLiA UNet1D-Large result instead uses the 80-epoch retuning protocol described in Section~\ref{app:unet_retuning}. Conditional DDPM-1D is trained using an $L_1$ noise-prediction loss with AdamW (learning rate $2\times10^{-4}$, batch size 128, and 80 epochs). All models are trained for fixed epoch horizons, and the checkpoint with the best validation performance is selected within each LOSO fold using only non-test subjects.

\subsection{Learning-rate selection for UNet1D-Large}
\label{app:unet_retuning}

We performed an additional tuning step for UNet1D-Large on PPG-DaLiA. The 13.2M-parameter model was trained for 80 epochs using three learning rates: $3\times10^{-4}$, $1\times10^{-4}$, and $3\times10^{-5}$. We used S6, S10, and S12 as development folds and selected the learning rate using validation L1 computed only from non-test subjects. The held-out subject in each development fold was not used for model selection. Table~\ref{tab:unet_lr_sweep} summarizes the learning-rate sweep.

\begin{table}[H]
\centering
\caption{\textbf{Learning-rate sweep for UNet1D-Large on PPG-DaLiA.} Validation L1 is computed using non-test subjects from the development folds.}
\label{tab:unet_lr_sweep}
\begin{tabular}{lc}
\toprule
\textbf{Learning rate} & \textbf{Validation L1} $\downarrow$ \\
\midrule
$3\times10^{-4}$ & \textbf{0.148} \\
$1\times10^{-4}$ & 0.165 \\
$3\times10^{-5}$ & 0.192 \\
\bottomrule
\end{tabular}
\end{table}

Based on the validation results, we fixed the learning rate to $3\times10^{-4}$ and reran all 15 LOSO folds. The final UNet1D-Large result on PPG-DaLiA is L1 0.138, RMSE 0.210, Pearson $r$ 0.925, R-peak F1 0.875, and HR error 6.95 bpm. These are the values reported in the main comparison table. Despite this tuning, UNet1D-Large does not improve over the smaller UNet1D under LOSO; we retain it as a capacity-matched control rather than assuming that increased model size should monotonically improve unseen-subject generalization.

\subsection{BeatFlow-ECG hyperparameters}
\label{app:hyper}

Table~\ref{tab:app_hparams} summarizes the BeatFlow-ECG configuration and the shared windowing settings used in the experiments. Baseline training settings are provided in Appendix~\ref{app:baseline_details}.

\begin{table}[!htbp]
\centering
\renewcommand{\arraystretch}{0.9}
\caption{\textbf{Main hyperparameters used in the reported experiments.}}
\label{tab:app_hparams}
\resizebox{0.75\textwidth}{!}{%
\begin{tabular}{lc}
\toprule
\textbf{Hyperparameter} & \textbf{Value} \\
\midrule
Window length & 1250 samples (10 s) \\
Stride & 625 samples (5 s) \\
Sampling rate & 125 Hz \\
BeatFlow-ECG transformer depth & 8 \\
Hidden dimension & 320 \\
Feedforward dimension & 1280 \\
Dropout & 0.0 \\
Optimizer & AdamW \\
Learning rate & $2\times10^{-4}$ \\
Batch size & 128 \\
Epochs & 80 \\
LR schedule & Cosine annealing \\
Flow-time sampling & $\mathrm{Beta}(0.5,0.5)$ \\
Minimum flow time & $10^{-3}$ \\
Primary loss & L1 velocity matching \\
STFT loss weight & 0.10 \\
STFT FFT size & 256 \\
Motion-weight coefficient $\alpha$ & 1.0 \\
Gradient clipping & 1.0 \\
Inference solver & Explicit Euler \\
Inference steps & 120 \\
Hardware & NVIDIA H100 80GB GPU \\
\bottomrule
\end{tabular}%
}
\end{table}
\FloatBarrier

\subsection{Ablation details}
\label{app:ablation_extended}

The main ablation results are reported in Table~\ref{tab:ablation_main}. Here, we provide additional details on the ablated configurations and how each experiment was constructed.

For the \emph{w/o Motion Conditioning} variant, the explicit IMU-derived motion descriptors are removed from the conditioning representation while the remaining architecture and training procedure are unchanged. L1 error increases from 0.062 to 0.078 and R-peak F1 decreases from 0.946 to 0.920, indicating that the motion-conditioning information contributes to reconstruction performance.

For the \emph{w/o Curriculum} variant, all training windows are used throughout training without the easy-to-hard motion schedule. L1 error increases to 0.071 and HR error to 5.35 bpm, indicating a benefit from the motion-based curriculum under the evaluated training setup.

For the \emph{w/o STFT Loss} variant, the spectral consistency term is removed while retaining the motion-weighted flow-matching objective. Performance decreases across all reported metrics, with L1 increasing from 0.062 to 0.069 and R-peak F1 decreasing from 0.946 to 0.926. This indicates that the STFT term provides additional benefit beyond the time-domain velocity-matching objective.

Finally, in the \emph{w/o Motion Weighting + STFT Loss} variant, both the motion-dependent weighting and the STFT objective are removed. This produces the largest degradation among the evaluated variants, increasing L1 to 0.084 and HR error to 5.85 bpm. Together with the individual ablations, these results show that the motion-aware training components and the STFT objective contribute to the performance of the full model.
\FloatBarrier

\subsection{Experimental workflow}
\label{app:commands}

The project uses a subject-independent LOSO pipeline in which each fold uses 14 subjects for training and validation and evaluates on the held-out subject. The workflow consists of: (i) preprocessing all subjects into paired wearable/ECG windows, (ii) constructing subject-level manifests and splits, (iii) training the selected model within each fold, and (iv) evaluating the validation-selected checkpoint on the held-out subject while saving aggregate metrics and per-window prediction artifacts.

\end{document}